\documentclass[10pt,twocolumn,letterpaper]{article}

\usepackage[pagenumbers]{cvpr}

\usepackage{booktabs}
\usepackage{colortbl}
\usepackage{multirow}
\usepackage{amsmath,amssymb}
\usepackage{algorithm}
\usepackage{algpseudocode}
\usepackage{xspace}
\usepackage{tikz}   
\usepackage[normalem]{ulem}
\usetikzlibrary{arrows.meta,positioning}

\newcommand{\method}{HSM\xspace}
\newcommand{\spotlight}{Spotlight\xspace}
\newcommand{\vlm}{Qwen3.5-9B\xspace}
\newcommand{\orch}{DeepSeek V4.1 Flash\xspace}
\newcommand{\judge}{gpt-4o-mini\xspace}

\newcommand{\Mpct}{M\%}
\definecolor{oodshade}{rgb}{0.90,0.96,0.90}
\newcommand{\sd}[1]{{\fontsize{4.5pt}{5pt}\selectfont$\pm$#1}}
\makeatletter
\renewcommand\paragraph{\@startsection{paragraph}{4}{\z@}{1.5ex \@plus .5ex \@minus .2ex}{-1em}{\normalfont\normalsize\bfseries}}
\makeatother

\usepackage{listings}
\lstdefinestyle{prompt}{basicstyle=\ttfamily\fontsize{7}{8.4}\selectfont, breaklines=true, breakatwhitespace=true, columns=fullflexible, keepspaces=true, frame=single, framesep=2pt, xleftmargin=3pt, xrightmargin=3pt, aboveskip=3pt, belowskip=6pt, showstringspaces=false, literate={→}{{$\rightarrow$}}1 {·}{{$\cdot$}}1}

\definecolor{cvprblue}{rgb}{0.21,0.49,0.74}
\usepackage[pagebackref,breaklinks,colorlinks,allcolors=cvprblue]{hyperref}

\def\confName{CVPR}
\def\confYear{2027}

\title{Optimising VLM Harnesses to Localise Errors in Generated Videos}

\author{Ankit Yadav$^{1}$ \quad Nikhil Cherian Kurian$^{1}$ \quad Arpit Garg$^{1}$ \quad Lingqiao Liu$^{1}$\\
$^{1}$Australian Institute for Machine Learning, University of Adelaide
}

\begin{document}
\maketitle

\begin{abstract}
    Identifying subtle errors in generated videos is an important step toward automated video generation, where errors must be detected to assess and refine generated clips. Recent work shows that pretrained vision-language models (VLMs) can localise such errors, but their effectiveness depends strongly on the workflow used: choices about which frames to inspect, which tools and model calls to make, and how to combine predictions can lead to substantially different results. We study VLM harnesses for video error localisation. Instead of relying on manually designed harnesses, we propose Harness Search and Merge (HSM), an algorithm that automatically constructs them by searching over executable evaluation programs. An LLM orchestrator refines candidate harnesses independently on random training subsets, records findings in a memory, and constructively merges complementary candidates. 
    We evaluate HSM on Spotlight, a recently proposed benchmark for localising errors in generated videos. HSM achieves $0.234 \pm 0.012$ F1, exceeding the strongest benchmark strategy ($0.173 \pm 0.016$), and retains its lead on an unseen video generator ($0.213 \pm 0.008$ versus $0.140 \pm 0.003$). On IPV-Bench, it achieves strong localisation performance with lower false-alarm rates than existing optimisation baselines. Together, these results show that searching over complete VLM harnesses can outperform existing evaluation strategies, transfer to an unseen video generator, and reduce false alarms on real videos.
\end{abstract}

\section{Introduction}
\label{sec:intro}
\begin{figure*}[t]
\centering
\begin{minipage}[c]{0.385\textwidth}
\centering
\resizebox{\linewidth}{!}{\begin{minipage}{270pt}\small
\algrenewcommand{\algorithmicrequire}{\textbf{Input:}}\algrenewcommand{\algorithmicensure}{\textbf{Output:}}%
\hrule height 0.8pt\vskip 2pt
\textbf{Discovered harness:} propose $\rightarrow$ refine $\rightarrow$ select.\par\vskip 2pt\hrule\vskip 2pt
\begin{algorithmic}[1]
\Require Video clip $x$ and generation prompt $p$
\Ensure Up to three predictions, each with a time segment and explanation

\Statex \textbf{1. Propose: find possible errors}
\State Scan 2.5\,s windows at 4\,fps; propose up to two errors per window.
\State Check the full clip at 2\,fps for prompt mismatches and physics errors.
\State Use RAFT, DINOv2, CLIP, and Grounding DINO to flag suspicious moments.
\State Ask the VLM to check up to five flagged moments at 8\,fps.
\State If no errors are proposed, rescan the full clip at 4\,fps.

\Statex \textbf{2. Refine: pinpoint each error}
\State Remove overlapping candidates, keeping shorter segments.
\State Recheck the eight shortest candidates at 8\,fps.
\State Update their explanations and limit segments to 0.6\,s.

\Statex \textbf{3. Select: rank and filter errors}
\State Add 2 points for errors from the full-clip prompt check,
\Statex \hspace{\algorithmicindent} 1 point for physics-related wording, and 1 point for being within
\Statex \hspace{\algorithmicindent} 0.5\,s of a tool-flagged moment.
\State Rank by score, preferring shorter segments when scores tie.
\State Remove vague flicker or morphing reports and overlapping duplicates.
\State \Return the top three predictions.
\end{algorithmic}
\vskip 2pt\hrule height 0.8pt
\end{minipage}}
\end{minipage}\hfill
\begin{minipage}[c]{0.595\textwidth}
\centering
\includegraphics[width=\linewidth]{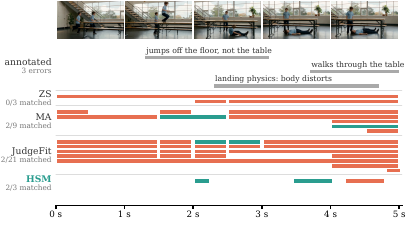}
\end{minipage}
\caption{\textbf{The harness discovered by \method and its predictions on a test clip.} \uline{\textit{Left}}: the best candidate from \cref{tab:results} (test F1: 0.247, best of three runs; mean: $0.234 \pm 0.012$), found by \cref{alg:hsm}. \uline{\textit{Right}}: a Seedance test clip (prompt 122), its three annotated errors (grey), and predictions from  three baselines (Qwen3.5-9B) and \method. Teal and orange denote matched and unmatched predictions, respectively. 
}
\label{fig:overview}
\end{figure*}

Video generation is moving toward increasingly automated systems that can create and refine content with limited human intervention~\cite{long2026vista,liu2024evalcrafter}. For such systems to be reliable, they need to assess their own outputs and identify what should be corrected. This is particularly important for generated videos, where a clip can appear plausible overall while containing subtle errors: an object may pass through a wall, a hand may gain an extra finger, or an object may unexpectedly change state. If these errors cannot be detected, a generation system has little basis for deciding whether or how to revise its output. Reliable error localisation is therefore an important capability for building more autonomous video generation and refinement systems.

Recent benchmarks have begun to study this problem directly. VBench~\cite{huang2024vbench} and EvalCrafter~\cite{liu2024evalcrafter} assess generated video quality along predefined dimensions, while Spotlight~\cite{chinchure2026spotlight} formulates error localisation as predicting temporal segments with short explanations. A prediction is considered correct only when it agrees with a human annotation in both temporal overlap and semantic content. Spotlight also demonstrates that pretrained vision-language models (VLMs) can identify and explain errors without task-specific fine-tuning. This is an encouraging direction: pretrained VLMs, including both open and closed models, offer a flexible basis for evaluating generated videos.

However, the performance of a pretrained VLM depends not only on the model, but also on how it is used. An evaluation workflow must decide which frames to inspect, how to organise model calls, whether and when to invoke vision tools, and how to combine the resulting predictions. Different workflows can lead to substantially different outcomes. For example, Spotlight’s strongest strategies decompose evaluation by error type or temporal window. This illustrates a broader challenge: a VLM’s ability to evaluate generated content depends on the harness, a complete workflow that connects the video, the model, tools, and prediction aggregation.

Most existing approaches improve parts of a predefined workflow, such as its instructions~\cite{agrawal2026gepa,yang2024large,yuksekgonul2024textgrad}, error taxonomy~\cite{cao2026each}, available tools~\cite{ye2026videoresearcher}, or evidence gathered for an individual input~\cite{videoargus,zhang2025evaluation}. The overall workflow, however, is generally specified by human designers. This raises the central question of our work: can an effective VLM harness be learned automatically from annotated examples? We formulate harness construction as a supervised-learning-style search problem. Given videos and human annotations, a search procedure proposes candidate harnesses, runs them on training examples, compares their predictions with the ground truth, and uses the resulting feedback to improve subsequent candidates. In this way, the system learns how to use a pretrained VLM for a task without fine-tuning the VLM itself.

Searching over complete harnesses introduces two challenges. First, evaluating each candidate requires executing it on videos, so testing many candidates on the full training set can be computationally expensive. Yet evaluating only a few candidates risks missing effective workflows. Second, a harness that performs well on one subset may not be robust across other videos, while different subsets may reveal different useful designs. A practical search should therefore use its evaluation budget efficiently and make use of complementary solutions discovered across the data.

We propose Harness Search and Merge (HSM) to address these challenges. HSM partitions the training data into random subsets and independently proposes candidate harnesses for each subset. Candidates are first evaluated and iteratively refined on their assigned subset, allowing an initially immature design to improve before further evaluation. After refinement, candidates are evaluated on the other subsets. These cross-subset results help identify promising harnesses and reveal complementary strengths. HSM then proposes merged harnesses that combine these strengths and evaluates them on the round’s training data. A held-out validation set ranks candidates across rounds, while its contents and scores remain hidden from the orchestrator. The VLM itself remains pretrained; human annotations guide the search over how the model is used rather than fine-tuning its parameters.

We evaluate HSM on Spotlight and IPV-Bench. On Spotlight, HSM with Qwen3.5-9B achieves $0.234 \pm 0.012$ detection F1, compared with $0.173 \pm 0.016$ for the strongest benchmark strategy, while producing approximately four predictions per clip. It maintains an advantage on an unseen video generator, achieving $0.213 \pm 0.008$ compared with $0.140 \pm 0.003$. HSM also transfers across VLMs: with Kimi-VL-A3B-Instruct, it achieves $0.219 \pm 0.028$ (\cref{tab:suppl:results_full}). 
On IPV-Bench, in-domain search more than doubles \method's F1$^*$ relative to direct transfer from \spotlight$,$ while reducing false alarms on real clips from nearly 100\% to 7.5\%, roughly half LoRA's rate and one-sixth of MIPROv2's.
These results show that automatically searching and merging complete harnesses can improve video error localisation across models, generators, and evaluation settings.
Our contributions are as follows:
\begin{enumerate}
    \item Problem formulation. We formulate VLM harness construction for video error localisation as a supervised search problem over complete executable evaluation workflows. We also show that directly optimising Spotlight’s reported score can encourage overprediction, motivating detection F1 as the search objective.
    \item Search algorithm. We introduce HSM, which combines 
    refinement on random training subsets, and constructive merging to discover and improve executable VLM harnesses.
    \item Evaluation. We evaluate HSM across Spotlight and IPV-Bench, including unseen-generator and cross-model settings. Our results show consistent gains over existing fixed strategies and optimisation baselines; ablations further show that independent refinement and constructive merging both contribute to performance. \footnote{Code will be released upon acceptance.}
\end{enumerate}

\section{Related Work}
\label{sec:related}

\paragraph{From quality scores to error localisation.}
VBench~\cite{huang2024vbench,zheng2503vbench} and EvalCrafter~\cite{liu2024evalcrafter} assess video generators along predefined quality dimensions without explicitly localising errors within individual clips. \spotlight~\cite{chinchure2026spotlight} instead evaluates error localisation using fixed prompting strategies as reference detectors, but its reported metric does not penalise unmatched predictions. We retain its task, annotations, and matching procedure while optimising detection F1 through search over executable detector harnesses.

\begin{figure}[t]
\centering
\includegraphics[width=\linewidth]{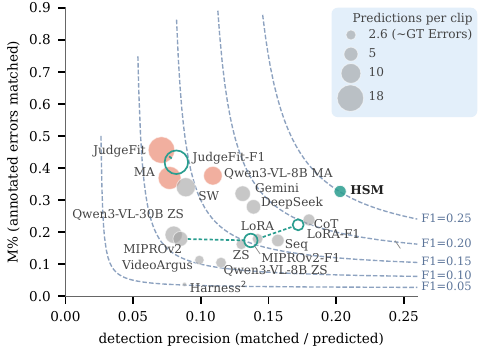}
\caption{\textbf{Precision--coverage trade-off on the test split.} \method and F1-based baselines (teal) balance precision and $\Mpct$, while the scalar's top three methods (orange) overpredict. Hollow markers denote F1-optimised baselines, with dotted lines connecting them to their scalar-optimised counterparts. Marker size denotes predictions per clip. Dashed iso-F1 curves approximate per-clip F1 within 0.02 (\cref{tab:results}). All methods: \cref{tab:suppl:results_full}.}

\label{fig:tradeoff}
\end{figure}

\paragraph{Agentic and rubric-based evaluation.}
A second line of work replaces a single judge prompt with a structured evaluation procedure. Evaluation Agent~\cite{zhang2025evaluation} decomposes evaluation
into sub-aspects, generates probe prompts, and invokes an evaluation toolkit to efficiently characterise a \emph{generator}; its open successor trains a local
planner to remove dependence on a proprietary model~\cite{tian2026open}. VideoArgus~\cite{videoargus} constructs an input-specific rubric without observing the
generated output. Each criterion specifies an evidence-gathering plan using VLM question answering and vision tools, including tracking, visual similarity, OCR,
depth, and flicker analysis, before a judge assigns a score and supporting rationale. VQQA~\cite{song2026vqqa} generates visual questions, answers them with a
VLM, and uses identified failures as a semantic gradient to refine the generation prompt. RewardVerse~\cite{tang2026rewardverse} introduces a generated rubric
between the query and scorer to stabilise pointwise rewards. Related generation frameworks incorporate agentic judges to inspect execution traces and final
videos~\cite{wei2026videoweaver} or rank candidates through tournaments~\cite{long2026vista}.

Although these methods adapt prompts, rubrics, or tool calls to individual inputs, their overall evaluation procedures remain manually specified. \method instead searches over the procedure itself (\cref{sec:problem}). Prior work also shows that a VLM verifier accepts most failures when execution traces report success~\cite{xu2026don}. To limit this bias, our judge receives only candidate predictions and reference annotations, without access to candidate code, tool outputs, or reasoning.

\paragraph{Optimising the evaluator itself.}
JudgeFit~\cite{cao2026each} uses an LLM to revise a VLM judge's physics-error taxonomy by merging, splitting, redefining, adding, or removing categories, retaining edits only when they improve calibrated agreement with human ratings on annotated data. Related prompt and program optimisers iteratively refine candidates using feedback. Self-Refine~\cite{madaan2023self} and Reflexion~\cite{shinn2023reflexion} retain verbal feedback from previous attempts; OPRO~\cite{yang2024large} conditions proposals on scored solutions; MIPROv2~\cite{opsahl2024optimizing} searches instructions and demonstrations for multi-stage language-model programs; TextGrad~\cite{yuksekgonul2024textgrad} propagates textual critiques through a computation graph; and GEPA~\cite{agrawal2026gepa} evolves prompts through reflection on execution traces and Pareto-based selection. AlphaEvolve~\cite{novikov2025alphaevolve} extends evolutionary search to code, selecting parents from a database of scored programs. 
ADAS~\cite{hu2025automated} represents agents as code and uses a meta-agent to iteratively generate new agents from a growing archive of prior candidates. AFlow~\cite{zhang2025aflow} searches over code-represented workflows using Monte Carlo Tree Search guided by execution feedback. OpenEvolve~\cite{openevolve} provides an open-source implementation of AlphaEvolve with a MAP-Elites program database distributed across island populations. 

VideoResearcher~\cite{ye2026videoresearcher} evolves the \emph{tools} available to a fixed long-video agent. Harness$^2$~\cite{xu2026harness} recursively edits an agent's harness (prompts, skills, scripts, and subagents) on unseen tasks without a verifier, comparing candidate executions against a reference and retaining evidence-backed lessons across tasks. Unlike these program-level methods, LoRA fine-tuning~\cite{hu2021lora} adapts the video model's weights through supervised training. \method searches over complete harnesses for supervised video error localisation, combining independent refinement on random training subsets and constructive merging.

\section{Method}
\label{sec:method}

We first revisit the video error localisation task and its evaluation metrics, and examine why fixed, manually designed harnesses leave the human--model performance gap identified in \cref{sec:intro} unresolved (\cref{sec:problem}). We then formulate harness design as a search problem and define the candidate space (\cref{sec:supervised_harness}). Finally, we present \method, describing its components and the overall search algorithm (\cref{sec:hsm}).

\subsection{Problem Formulation}
\label{sec:problem}

Let $v$ be a generated video, let $q$ be the text prompt used to
generate it, and let $L$ be the duration of $v$. Our goal is to
identify and temporally localise errors in $v$. We represent each
annotated error as a pair $(\tau,e)$, where $\tau=[t_s,t_e]$ is its
temporal interval, with $0 \leq t_s < t_e \leq L$, and $e$ is a
textual explanation. The annotation set for $v$ is
\[
Y = \{(\tau_j,e_j)\}_{j=1}^{K},
\]
where $K$ is the number of annotated errors.

We call an executable error-detection procedure a \emph{harness},
denoted by $h$. In other words, a harness is a set of external tools and
workflow steps built around a VLM to guide how it inspects video
evidence, detects errors, and turns its findings into temporally
localised predictions. Given $v$ and $q$, a harness returns a set of
predicted errors:
\[
\hat{Y}(v,q;h)
=
\{(\hat{\tau}_j,\hat{e}_j)\}_{j=1}^{\hat{K}},
\]
where $\hat{\tau}_j$ is a predicted temporal interval,
$\hat{e}_j$ is its explanation, and $\hat{K}$ is the number of
predictions made by $h$.

A harness coordinates a fixed pretrained vision-language model (VLM)
$\phi$ and a fixed set of available vision tools $\mathcal{T}$.
It specifies how video evidence is gathered and processed, including
which frames or temporal windows to inspect, how to prompt $\phi$,
which tools in $\mathcal{T}$ to invoke and in what order, and how to
combine their outputs into predictions. Thus, the harness defines the
error-detection workflow, while $\phi$ and $\mathcal{T}$ provide its
fixed model and tool components.

Figure~\ref{fig:overview} illustrates one harness discovered by our
method. It proposes candidate errors through temporal-window scanning,
a prompt check, and tool-based cues. It then refines the candidate
segments, for example by merging overlapping predictions and tightening
their temporal boundaries, before selecting a small set of final
predictions. Other harnesses may use different stages, tools, or call
sequences; these choices are part of the harness being learned, rather
than a fixed design imposed by a human expert.

\subsection{Supervised Harness Optimisation}
\label{sec:supervised_harness}

We learn a harness from labelled examples. Let
\[
\mathcal{D}_{\mathrm{tr}}
=
\{(v_i,q_i,Y_i)\}_{i=1}^{N_{\mathrm{tr}}}
\]
be the training set, where $v_i$ is a generated video, $q_i$ is its
generation prompt, and $Y_i$ is the set of human-annotated errors.
Let $\mathcal{H}$ denote the space of executable harnesses, and let
$M(h;\mathcal{D}_{\mathrm{tr}})$ measure the agreement between the
predictions of harness $h$ and the annotations in
$\mathcal{D}_{\mathrm{tr}}$. Our objective is
\[
h^\star =
\arg\max_{h \in \mathcal{H}}
M(h;\mathcal{D}_{\mathrm{tr}}).
\]
We use a separate labelled validation set $\mathcal{D}_{\mathrm{val}}$
to rank candidate harnesses during search and reserve the test set for
final evaluation.

Optimising a harness is non-differentiable: it is an executable
workflow that may select frames, call a VLM or external tools, and
process their outputs. We therefore use an iterative
\emph{propose--execute--revise} process. An orchestrator proposes a
harness, which is executed on labelled videos; the resulting
predictions and annotation-based feedback guide subsequent proposals.
Previous candidates and their outcomes are retained in a search memory
and archive, allowing later proposals to build on earlier search
experience.

In our implementation, $M$ is detection F1, computed for each video and
then averaged. Following Spotlight~\cite{chinchure2026spotlight}, we match predicted
errors to annotations using the Hungarian algorithm. For each
prediction--annotation pair, we average its temporal-overlap score and
semantic-similarity score; an LLM judge assesses semantic similarity by
comparing the predicted reason with the ground-truth explanation. We
retain assignments whose combined score is at least $0.7$.

We choose detection F1 because Spotlight's original metric does not
penalise unmatched predictions. A harness can therefore obtain a high
score by reporting many candidate errors, including false alarms.
Detection F1 accounts for both missed annotations and unmatched
predictions. Let $m_i$ be the number of retained matches for video
$v_i$, $K_i$ the number of annotated errors, and $\hat{K}_i$ the number
of predictions. We define detection precision $P_i$ and recall $R_i$ as
\[
P_i = \frac{m_i}{\hat{K}_i},
\qquad
R_i = \frac{m_i}{K_i},
\]
and compute per-video detection F1 as
\[
F_{1,i}^{\mathrm{det}}
=
\frac{2P_iR_i}{P_i+R_i}.
\]
The score is zero for videos with annotations but no predictions, or
predictions but no annotations; videos with neither are excluded from
the average. We use mean per-video detection F1 as the search
objective. Further details of the matching protocol are provided in
Section~\ref{sec:results:setup}.

\subsection{Harness Search and Merge}
\label{sec:hsm}

Searching over complete harnesses requires repeatedly executing
candidates on labelled videos. Testing every proposed design on the
full training set would limit the number of possibilities explored.
HSM therefore first proposes and refines candidates on small subsets,
allowing immature designs to improve before spending the budget on
broader testing. However, performance on a small subset can be noisy.
We consequently retain multiple refined solutions, assess them across
subsets, and combine candidates with complementary strengths. This
design supports efficient exploration while seeking harnesses that
perform robustly across videos.

Algorithm~\ref{alg:hsm} summarises the procedure. For compactness, let
$T=\mathcal{D}_{\mathrm{tr}}$ and
$V=\mathcal{D}_{\mathrm{val}}$ denote the training and validation sets.
HSM runs for $n$ rounds, maintaining a search memory $\mathcal{L}$.
Each round partitions $T$ into $m$ random, disjoint subsets
$\{S_j\}_{j=1}^{m}$. For each subset, the orchestrator proposes $k$
candidate harnesses using the previous search records and refines each
candidate independently. The resulting finalists are evaluated outside
their refinement subsets and used to propose merged harnesses.
Both selected finalists and successful merges enter the search memory,
so useful individual solutions remain available alongside their
combinations.

\begin{algorithm}[t]
\footnotesize
\caption{Harness Search and Merge (HSM)}
\label{alg:hsm}
\begin{algorithmic}[1]
\Require Training set $T$, validation set $V$
\Require Rounds $n$, subsets $m$, lineages $k$, refinement steps $s$
\Require Merge iterations $u$, proposals $q_m$, retained counts $a,b$
\Ensure Selected harness $h^\star$

\State $\mathcal{L} \gets \emptyset$
\For{$r=1,\ldots,n$}
    \State $\{S_j\}_{j=1}^{m} \gets \Call{Partition}{T,m}$
    \State $\mathcal{X}_r,\mathcal{M}_r \gets \emptyset,\emptyset$
    \For{$j=1,\ldots,m$}
        \State $\mathcal{H}_j \gets \Call{Propose}{k,\mathcal{L},S_j}$
        \ForAll{$h \in \mathcal{H}_j$}
            \State $h \gets \Call{Refine}{h,S_j,s}$
            \State $x_h \gets \Call{CrossEval}{h,T\setminus S_j}$
            \State $\mathcal{X}_r \gets \mathcal{X}_r \cup \{(h,x_h)\}$
        \EndFor
    \EndFor
    \For{$t=1,\ldots,u$}
        \State $\mathcal{G}_t
            \gets \Call{MergePropose}{q_m,\mathcal{X}_r,\mathcal{M}_r}$
        \ForAll{$g \in \mathcal{G}_t$}
            \If{$\Call{AcceptMerge}{g,T}$}
                \State $\mathcal{M}_r \gets \mathcal{M}_r \cup \{g\}$
            \EndIf
        \EndFor
    \EndFor
    \State $\mathcal{E}_r \gets \Call{TopCross}{\mathcal{X}_r,a}$
    \State $\mathcal{B}_r \gets \Call{TopVal}{\mathcal{M}_r,V,b}$
    \State $\mathcal{L}
        \gets \Call{Update}{\mathcal{L},\mathcal{E}_r\cup\mathcal{B}_r,V}$
\EndFor
\State \Return $\Call{Best}{\mathcal{L}}$
\end{algorithmic}
\end{algorithm}

\paragraph{Independent refinement.}
Each initial harness forms an independent lineage and undergoes $s$
refinement steps on its assigned subset. At each step, the orchestrator
critiques the current harness using its predictions and annotation-based
feedback, then implements a revision. The revision replaces the current
harness only if it improves the objective $M$ without reducing the
number of matched annotations. Each lineage accesses its own refinement
history, including unsuccessful revisions.

Because scores on a small subset can be noisy, we retain multiple best
accepted harnesses from each lineage rather than selecting a single
winner for the subset. These multiple finalists preserve promising
alternatives for broader testing and subsequent merging.

\paragraph{Cross-subset evaluation.}
Each finalist $h$ is evaluated on training videos outside its
refinement subset, producing a record $x_h$ containing its score and
prediction-level results. The collection
$\mathcal{X}_r=\{(h,x_h)\}$ provides evidence for comparing harnesses
developed on different subsets. Since these videos were not used to
refine the corresponding candidate, the results provide a more reliable
basis for selection than its refinement score alone. They also reveal
which annotated errors are detected by different candidates.

\paragraph{Constructive merging.}
The orchestrator receives candidate summaries and a complementarity
table identifying which annotated errors each finalist detects.
It also receives the full code of the three finalists with the highest
cross-subset scores. Over $u$ iterations, it proposes $q_m$ merged
harnesses per iteration, each specifying at least two parents and their
intended contributions. Parents can be finalists or earlier merges
from the same round.

Each merged harness is executed on the round's training data. It is
retained only if its matched-error count and objective $M$ are both
at least as high as those of its best parent, where the best parent
is determined by $M$ on the same data. This check guards against
losing the strengths of the original harnesses when combining them.

\paragraph{Selection and search memory.}
At the end of each round, $\Call{TopCross}{\mathcal{X}_r,a}$ selects
the $a$ finalists with the highest cross-subset scores.
$\Call{TopVal}{\mathcal{M}_r,V,b}$ evaluates retained merges on $V$
and selects the best $b$. Both groups enter the search memory
$\mathcal{L}$, where candidates are ordered by validation performance.
Thus, subsequent rounds can build on strong individual harnesses as
well as merged solutions.

The memory stores compact candidate summaries, while an archive retains
code, predictions, and feedback for selective retrieval. Validation
examples and numerical scores are hidden from the orchestrator;
only the candidate ordering is exposed. After $n$ rounds, HSM returns
the harness with the highest validation score.

\section{Experiments and Results}
\label{sec:results}

\subsection{Setup}
\label{sec:results:setup}
\paragraph{\spotlight Bench.}
\spotlight contains 600 clips from 200 prompts rendered by three generators (Seedance, Veo~3, and LTX-2), with 1{,}604 annotated errors. We reserve LTX-2 for \emph{out-of-distribution} (OOD) evaluation and partition the remaining clips by prompt into a \emph{test split} of 40 prompts (80 clips, 205 errors) and an \emph{optimisation pool} of 160 prompts (320 clips), preventing test-prompt overlap with training under either generator. From this pool, we sample by prompt a 50-clip \emph{\spotlight-50 subset} (25 prompts), used for HSM experiments unless otherwise stated, and a 200-clip training set (100 prompts) for full-pool runs. A stratified validation set $\mathcal{V}$ (24 clips; 80 for full-pool runs; \cref{sec:hsm}) is drawn from pool prompts excluded from the corresponding training set. The OOD test comprises all 200 LTX-2 clips (580 errors), sharing 160 prompts with the optimisation pool but using a generator unseen during search, training, or prompt optimisation. All splits are sampled once with seed 8 and shared across experiments. Test and OOD data are accessed only for final evaluation of the optimised harness.

\paragraph{IPV-Bench.}
IPV-Bench~\cite{bai2025impossible} contains 902 generated clips of impossible events and 650 real clips. Each generated clip has a free-text answer with three alternative phrasings but no temporal error annotations. We evaluate two settings: direct OOD transfer of \spotlight-optimised harnesses and in-domain optimisation on IPV-Bench. Generated clips are partitioned by generator family into 400 training and 499 test clips. We hold out 400 real clips for false-alarm evaluation and sample real training clips from the remaining 250 real clips. For in-domain optimisation, \method uses 50 impossible clips, while LoRA and MIPROv2 use all 400, with each method additionally using real clips sampled from the remaining 250-clip pool (\cref{tab:suppl:ipv_subset}). Methods receive only the video, without the generation prompt, which would reveal the impossible event. We retain \spotlight's judge and matching rule but omit temporal overlap: $P$: a prediction matches if its semantic score $S$, averaged over all four reference phrasings, is at least $0.7$. $\Mpct$ is the fraction of impossible test clips with a matched prediction. Every prediction on a real clip is a false alarm; we report the fraction of real clips receiving at least one prediction (\cref{tab:ipv}; details in \cref{sec:suppl:ipv}).

\paragraph{Scoring.}
We report the metrics defined in \cref{sec:problem}: \spotlight's scalar, $\Mpct$, precision, and detection F1, computed per clip and averaged across clips, alongside predictions per clip. For brevity, we refer to detection F1 as F1 from here on. All systems use the same judge, \judge. On IPV-Bench, each impossible clip contains one annotated event and each real clip contains none. We therefore report F1$^*$ over all test clips rather than per-clip F1, with recall given by $\Mpct$ and precision by the fraction of all predictions matched to their clip's event (\cref{sec:suppl:ipv_f1}). Unless otherwise stated, all methods use \vlm as the VLM and \orch for \method orchestration and any baseline optimisation requiring an LLM. Tables and figures numbered with an S are in the supplementary material.

\label{sec:method:instantiation}
\method uses the configuration in \cref{tab:suppl:hparams}. The main results, ablations, grid, and sensitivity analysis use the \spotlight-50 budget (two rounds over 50 training clips, with 24 validation clips). \method-Full is the full-pool run (three rounds over 200 training clips), reported alongside it in \cref{tab:results,tab:results_vlm}. In rows marked Qwen-orch, \vlm replaces both the orchestrator \orch and the in-loop judge \judge, with a large thinking budget (\cref{sec:suppl:qwen_orch}). For training and optimisation, LoRA and MIPROv2 use the full 320-clip pool, with LoRA reserving 40 clips for checkpoint selection. \method results are reported as mean $\pm$ standard deviation over three runs with a fixed configuration and seed, varying only the orchestrator's sampling.

\subsection{Baselines}
\label{sec:results:baselines}
\Cref{tab:baselines} lists the baselines, what each adapts, and the data it uses; all are scored under the same protocol (implementation details in \cref{sec:suppl:baselines}). To assess dependence on the VLM and orchestrator, we repeat the search with three additional VLMs (\cref{tab:suppl:vlms}) and two additional orchestrators (\cref{tab:suppl:grid}).

\begin{table}[t]
\centering
\footnotesize
\setlength{\tabcolsep}{3pt}
\resizebox{\linewidth}{!}{%
\begin{tabular}{llll}
\toprule
ID & Baseline & Adapted & Data used (objective) \\
\midrule
A1--A2 & \spotlight prompts: ZS, MA, Seq, SW, CoT~\cite{chinchure2026spotlight} & none (released prompts) & none \\
A6 / A7 & ZS, MA with Qwen3-VL-8B; ZS with Qwen3-VL-30B & none (released prompts) & none \\
A8 / A8b & Gemini~2.5~Pro / \orch, ZS on frames & none (released prompt) & none \\
A8c / A8d & GLM~5.3~Flash / MiniMax~M3.1~Flash, ZS on frames & none (released prompt) & none \\
A3 & LoRA~\cite{hu2021lora} & VLM weights & 280 pool clips (supervised labels) \\
A4 & MIPROv2~\cite{opsahl2024optimizing} & ZS prompt & 320 pool clips (scalar) \\
A4-F1 & MIPROv2-F1 & ZS prompt & 320 pool clips (F1) \\
A11 & JudgeFit-loc~\cite{cao2026each} & MA category list & 320 pool clips (scalar) \\
A11-F1 & JudgeFit-loc-F1 & MA category list & 320 pool clips (F1) \\
A12 & VideoArgus-loc~\cite{videoargus} & per-prompt rubric & none \\
A13 & Harness$^2$~\cite{xu2026harness} & harness & \spotlight-50 (no verifier) \\
\midrule
--- & \method (ours) & harness code & \spotlight-50 (F1) \\
--- & \method-Full & harness code & 200 pool clips (F1) \\
\bottomrule
\end{tabular}}
\caption{\textbf{Baselines.} All methods use \vlm except A6--A8d, which use the listed VLM, optimizing either \spotlight's reported score ($(M+\Mpct)/2$) or F1 (\cref{sec:supervised_harness}).}
\label{tab:baselines}
\end{table}

\subsection{Results}
\label{sec:results:main}

\begin{table}[t]
\centering
\small
\setlength{\tabcolsep}{3pt}\renewcommand{\arraystretch}{0.85}
\resizebox{\linewidth}{!}{%
\begin{tabular}[t]{lc>{\columncolor{oodshade}}cc>{\columncolor{oodshade}}c}
\toprule
 & \multicolumn{4}{c}{\vlm} \\
\cmidrule(lr){2-5}
 & \multicolumn{2}{c}{F1\,$\uparrow$} & \multicolumn{2}{c}{pred/clip} \\
\cmidrule(lr){2-3}\cmidrule(lr){4-5}
Method & Test & OOD & Test & OOD \\
\midrule
ZS (A1) & .141 & .125 & 3.0 & 3.3 \\
MA (A2) & .130 & .139 & 13.8 & 14.3 \\
Seq (A2) & .155 & .122 & 3.9 & 3.7 \\
SW (A2) & .132 & .133 & 9.2 & 10.8 \\
CoT (A2) & .173 & .140 & 3.3 & 3.4 \\
\addlinespace[2pt]
LoRA (A3) & .136 & .148 & 3.3 & 3.4 \\
LoRA-F1 (A3) & .154 & .159 & 3.1 & 3.2 \\
MIPROv2 (A4) & .120 & .122 & 4.2 & 4.8 \\
MIPROv2-F1 (A4) & .141 & .130 & 3.6 & 3.7 \\
\addlinespace[2pt]
JudgeFit (A11) & .121 & .111 & 18.3 & 20.5 \\
JudgeFit-F1 (A11) & .134 & .141 & 11.8 & 13.0 \\
VideoArgus (A12) & .087 & .069 & 2.5 & 2.4 \\
Harness$^2$ (A13) & .087 & .074 & 1.9 & 2.2 \\
\cmidrule{1-5}
\method (Deepseek-orch) & .234 & .213 & 3.9 & 4.0 \\
\method-Full (Deepseek-orch) & .241 & .239 & 3.5 & 3.6 \\
\method (QwenThinking-orch) & .284 & .290 & 5.4 & 6.6 \\
\method-Full (QwenThinking-orch) & \textbf{.340} & \textbf{.321} & 4.5 & 5.7 \\
\bottomrule
\end{tabular}%
\hspace{8pt}%
\begin{tabular}[t]{lc>{\columncolor{oodshade}}cc>{\columncolor{oodshade}}c}
\toprule
 & \multicolumn{4}{c}{Other VLMs} \\
\cmidrule(lr){2-5}
 & \multicolumn{2}{c}{F1\,$\uparrow$} & \multicolumn{2}{c}{pred/clip} \\
\cmidrule(lr){2-3}\cmidrule(lr){4-5}
Method & Test & OOD & Test & OOD \\
\midrule
\multicolumn{5}{c}{Qwen3-VL-8B} \\
ZS (A6) & .119 & .132 & 3.1 & 3.7 \\
MA (A6) & .164 & .163 & 8.8 & 9.8 \\
\midrule
\multicolumn{5}{c}{Qwen3-VL-30B} \\
ZS (A7) & .107 & .135 & 8.2 & 8.1 \\
\midrule
\multicolumn{5}{c}{Gemini-2.5-Pro} \\
ZS (A8) & .167 & .160 & 5.8 & 6.2 \\
\midrule
\multicolumn{5}{c}{DeepSeek-V4.1} \\
ZS (A8b) & .175 & .157 & 5.7 & 6.5 \\
\midrule
\multicolumn{5}{c}{GLM-5.3-Flash} \\
ZS (A8c) & .200 & .191 & 5.1 & 5.4 \\
\midrule
\multicolumn{5}{c}{MiniMax-M3.1-Flash} \\
ZS (A8d) & .187 & .185 & 7.1 & 7.0 \\
\bottomrule
\end{tabular}}
\caption{\textbf{Video-error localisation on \spotlight.}
Evaluation uses 80 test and 200 OOD LTX-2 clips
(\cref{sec:results:setup}). The left panel compares baselines
and \method using \vlm (\cref{sec:results:baselines});
the right compares zero-shot (ZS)
baselines using other VLMs. F1 and predictions per clip
(pred/clip) are averaged over three runs (\cref{tab:suppl:sd}); bold indicates
the best F1. Full metrics in
\cref{tab:suppl:results_full,tab:results_vlm};
}
\label{tab:results}
\end{table}

\paragraph{Results on the \spotlight test split (\cref{tab:results,tab:suppl:results_full}).}
\method achieves an F1 of $0.234 \pm 0.012$ and precision of $0.203 \pm 0.028$, with approximately four predictions per clip. Recall-oriented strategies (MA, SW, JudgeFit-loc) improve \spotlight's scalar and $\Mpct$ by producing 9--18 predictions per clip at the cost of precision (\cref{sec:problem}). Despite using 280--320 training clips, compared with 50 for \method, LoRA and MIPROv2 underperform the benchmark's CoT prompt ($0.136 \pm 0.035$ and
$0.120 \pm 0.013$ versus $0.173 \pm 0.016$ F1), while zero-shot frontier models achieve 0.167--0.200 F1. Among adapted agentic evaluators, JudgeFit-loc inherits MA's overprediction (18 predictions per clip), whereas VideoArgus-loc predicts sparingly (2.5 per clip) but matches few errors ($\Mpct=0.114$). Harness$^2$ edits the same harness without labels yet underperforms its zero-shot starting point (0.087 versus 0.141 F1), producing 156 predictions across 80 clips.

With Kimi-VL-A3B-Instruct as the VLM, baseline F1 ranges from 0.018 to 0.145, while \method achieves $0.219 \pm 0.028$ across three runs, with the highest precision (\cref{tab:suppl:results_full}). Optimising F1 rather than the scalar also reduces overprediction: JudgeFit-F1 produces approximately 36--37\,\% fewer predictions while gaining 0.013--0.030 F1, and MIPROv2-F1 produces 16--24\,\% fewer predictions while gaining 0.008--0.021 F1 across the test and OOD splits.

With a similar prediction count to CoT (309 versus 266), \method matches 50\,\% more annotated errors ($\Mpct$: $0.327 \pm 0.024$ versus $0.218 \pm 0.022$) at higher precision ($0.203 \pm 0.028$ versus $0.171 \pm 0.013$). A paired bootstrap over the 80 test clips (\cref{tab:suppl:bootstrap_ci}) gives 95\,\% intervals excluding zero for the difference from every baseline in that table, including CoT ($+0.061$, $+0.016$ to $+0.106$).

The Qwen-orch configuration, using \vlm as orchestrator and in-loop judge with thinking enabled and a larger output budget (\cref{sec:suppl:qwen_orch}), further raises test F1 from 0.234 to 0.284 on \spotlight-50 and from 0.241 to 0.340 on the full pool (OOD: 0.213 to 0.290 and 0.239 to 0.321). These results show that a small open-weight model with a larger thinking budget achieves competitive performance.

\paragraph{The discovered harness (\cref{fig:overview}).}
The harness discovered by \cref{alg:hsm} has three stages. First, it proposes errors using 2.5\,s windows at 4\,fps (up to two per window), a whole-clip prompt and physics check at 2\,fps, and RAFT, DINOv2, CLIP, and Grounding DINO cues (up to five moments verified at 8\,fps by the VLM). Second, it removes overlaps, favouring shorter segments, and re-localises the eight shortest at 8\,fps, capping segments at 0.6\,s. Finally, it assigns two points for prompt-check origin, one for physics wording, and one for proximity within 0.5\,s of a tool cue. It filters flicker or morphing reports naming no prompt object and near-duplicates, then returns the top three predictions to control overprediction.

\paragraph{Results on the \spotlight OOD split (\cref{tab:results,tab:suppl:results_full}).}
On the LTX-2 OOD set, whose generator was excluded from harness optimisation and baseline training, \method retains the highest F1 and precision despite a decline in $\Mpct$ from 0.327 to 0.284. Its F1 reaches $0.213 \pm 0.008$  versus $0.191 \pm 0.008$ for 
the strongest baseline, GLM 5.3 Flash zero-shot (LoRA-F1: $0.159 \pm 0.017$).
Three of the top five in-distribution baselines (the GLM 5.3 Flash, MiniMax M3.1 Flash and Gemini 2.5 Pro zero-shot models) remain in the top five. CoT, the strongest fixed prompt in distribution, falls to 0.140 F1 and, alongside Seq, exhibits the largest precision decline. Recall-oriented strategies with \vlm produce 10--21 predictions per clip and have the lowest precision on this split.
On the 40 LTX-2 clips whose prompts are also unseen (the test prompts; OOD-40 in \cref{tab:suppl:results_full}), \method reaches an F1 of $0.217 \pm 0.028$, against 
$0.172 \pm 0.026$ for the strongest baseline there (CoT).

\begin{table}[t]
\centering
\footnotesize
\setlength{\tabcolsep}{3pt}
\resizebox{\ifdim\width>\linewidth\linewidth\else\width\fi}{!}{%
\begin{tabular}{lcccccc}
\toprule
 & \multicolumn{3}{c}{Test F1} & \multicolumn{3}{c}{OOD (LTX-2) F1} \\
\cmidrule(lr){2-4}\cmidrule(lr){5-7}
VLM & Zero-shot & MA & \method & Zero-shot & MA & \method \\
\midrule
\vlm                 & .141 & .130 & .234 & .125 & .139 & .213 \\
Kimi-VL-A3B-Instruct & .066 & .088 & .219 & .063 & .097 & .202 \\
InternVL3.5-8B       & .117 & .119 & .175 & .118 & .129 & .180 \\
Gemma 4 E4B          & .132 & .105 & .217 & .154 & .126 & .186 \\
\bottomrule
\end{tabular}}
\caption{\textbf{Across VLMs.} \spotlight's ZS, MA baselines and \method (\orch orchestrator, \spotlight-50 budget) evaluated with different VLMs using \spotlight's settings, and per-clip F1 as in \cref{tab:results}. All results are means over three runs and standard deviations are reported in \cref{tab:suppl:sd}.}
\label{tab:suppl:vlms}
\end{table}

\begin{table*}[t]
\centering
\small
\setlength{\tabcolsep}{5pt}
\begin{tabular}{lcccc}
\toprule
Orchestrator $\downarrow$ / VLM $\rightarrow$ & \vlm & Kimi-VL-A3B-Instruct & InternVL3.5-8B & Gemma 4 E4B \\
\midrule
\orch                & 0.234 / 0.213 & 0.219 / 0.202 & 0.175 / 0.180 & 0.217 / 0.186 \\
GLM 5.3 Flash        & 0.164 / 0.168 & 0.215 / 0.203 & 0.164 / 0.144 & 0.157 / 0.149 \\
MiniMax M3.1 Flash   & 0.170 / 0.175 & 0.178 / 0.182 & 0.166 / 0.176 & 0.134 / 0.135 \\
\bottomrule
\end{tabular}
\caption{\textbf{Effect of orchestrator and VLM selection on \method.} Each cell reports test / OOD (LTX-2) F1. Every run uses
the \spotlight-50 budget (50 clips, two rounds, 144 harnesses) and the \judge judge across orchestrator and VLM grid. Cells are means over three runs. Standard deviations are reported in \cref{tab:suppl:sd}.}
\label{tab:suppl:grid}
\end{table*}

\paragraph{Performance versus cost (\cref{tab:suppl:cost}).}
\method achieves higher test/OOD F1 (0.234/0.213) than MIPROv2 (0.120/0.122), LoRA (0.136/0.148), and Harness$^2$ (0.087/0.074). It uses 50 training clips plus 24 for validation, compared with 320 for MIPROv2 and 280 training clips plus 40 for checkpoint selection for LoRA; Harness$^2$ uses the 50 \spotlight-50 clips without labels. A \method search costs approximately 14 GPU-hours and US\$4.6 in API calls, versus 5 GPU-hours and US\$2.5 for MIPROv2, 13 GPU-hours with no API cost for LoRA, and 4 GPU-hours and US\$8.4 for Harness$^2$. At inference, selected harnesses average 21 VLM calls and 84k prompt tokens per clip, approximately twice MA's usage and 1.6 times JudgeFit's, while producing at most one-third as many predictions with at least \textbf{2.2} times the precision (\cref{tab:suppl:cost,tab:suppl:results_full}).

\paragraph{In-domain and OOD results on IPV-Bench (\cref{tab:ipv,tab:suppl:ipv_full}).}
Under OOD transfer, \spotlight-optimised methods (LoRA, MIPROv2, and \method) flag 99.5--100\,\% of real clips, consistent with optimisation exclusively on clips containing errors. In-domain optimisation includes real clips and penalises false alarms. Across three runs, \method achieves $0.520 \pm 0.053$ F1$^*$, between LoRA ($0.588 \pm 0.014$) and MIPROv2 ($0.439 \pm 0.012$). The gap to LoRA is consistent with more frequent abstention: \method returns no prediction on 17\,\% of impossible clips, versus 1\,\% for LoRA. LoRA's explanations average 19 words, matching the ground-truth annotations and suggesting adaptation to their style, whereas \method's average 29 words; nevertheless, match rates on answered clips remain similar (63\,\% versus 59\,\%; \cref{sec:suppl:ipv}). \method achieves the lowest false-alarm rate on real clips ($7.5 \pm 3.3\,\%$), compared with $15.2 \pm 4.0\,\%$ for LoRA and $47.8 \pm 21.1\,\%$ for MIPROv2 (\cref{tab:ipv}).

\begin{table}[t]
\centering
\small
\setlength{\tabcolsep}{4pt}
\resizebox{\linewidth}{!}{%
\begin{tabular}{lcc}
\toprule
 & all test clips (899) & real clips (400) \\
\cmidrule(lr){2-2}\cmidrule(lr){3-3}
Method (\vlm)               & F1$^*$\,$\uparrow$ & false alarms\,$\downarrow$ \\
\midrule
\multicolumn{3}{l}{\emph{Fixed prompts}} \\
ZS (A1)                     & .231 & 99.0\,\% \\
MA (A2)                     & .291 & 72.0\,\% \\
Seq (A2)                    & .204 & 99.5\,\% \\
SW (A2)                     & .183 & 100\,\% \\
CoT (A2)                    & .162 & 86.5\,\% \\
Qwen3-VL-30B ZS (A7)        & .200 & 100\,\% \\
\midrule
\multicolumn{3}{l}{\emph{Transferred from \spotlight, run as is}} \\
LoRA (A3)                   & .168 & 100\,\% \\
MIPROv2 (A4)                & .198 & 100\,\% \\
\method (Ours)              & .227 & 99.5\,\% \\
\midrule
\multicolumn{3}{l}{\emph{Optimised on the IPV pool}} \\
LoRA (A3)                   & \textbf{.588}\sd{.014} & 15.2\sd{4.0}\,\% \\
MIPROv2 (A4)                & .439\sd{.012} & 47.8\sd{21.1}\,\% \\
\method (Ours)              & .520\sd{.053} & \textbf{7.5}\sd{3.3}\,\% \\
\bottomrule
\end{tabular}}
\caption{\textbf{IPV-Bench.} We report F1$^*$ on test-set
and false-alarm rate (\cref{sec:results:setup}).
\method, LoRA, and MIPROv2 are optimised on IPV training set,
with results averaged over three runs ($\pm$ SD).
Bold indicates the best per column, and additional metrics
in \cref{tab:suppl:ipv_full}.}
\label{tab:ipv}
\end{table}

\subsection{Ablation: search design}
\label{sec:results:ablation}

\begin{table}[t]
\centering
\footnotesize
\setlength{\tabcolsep}{4pt}
\newcommand{\abltick}{\textcolor[rgb]{0.0,0.55,0.2}{\checkmark}}
\resizebox{\linewidth}{!}{%
\begin{tabular}{lcccc}
\toprule
\method search design & Single split & Multi split & Merge & Test F1\,$\uparrow$ \\
\midrule
Single split, no merge & \abltick & & & 0.166\,$\pm$\,0.031 \\
Multi split, no merge & & \abltick & & 0.191\,$\pm$\,0.077 \\
\midrule
Multi split and merge (\method) & & \abltick & \abltick & \textbf{0.234}\,$\pm$\,0.012 \\
\bottomrule
\end{tabular}}
\caption{\textbf{Search design ablation.}
We evaluate \method components on \spotlight using
the \spotlight-50 subset, \orch as orchestrator, and
\vlm as VLM (\cref{sec:results:ablation}).
Test F1 is averaged over three runs ($\pm$ standard deviation;
\cref{tab:suppl:sd}).}
\label{tab:ablation}
\end{table}

All ablation searches share the \spotlight-50 subset, \orch, \vlm, and seed, with a comparable budget of 120--160 harnesses each (144 for \method; \cref{sec:suppl:budgets}), and we report the test F1 of the candidate selected on the held-out validation set. \Cref{tab:ablation} builds up the search design in two steps. The single-split variant searches all training clips at once, proposing new candidates at each iteration, and reaches $0.166 \pm 0.031$. Searching random, disjoint training subsets independently, each with persistent lineages refined through critique and rewriting, raises F1 to $0.191 \pm 0.077$, but with high variability across runs. Adding constructive merging across subsets yields $0.234 \pm 0.012$, the best mean and the lowest variability; on the OOD set the last step improves F1 from $0.199 \pm 0.053$ to $0.213 \pm 0.008$.

\noindent\textbf{Is the VLM essential?} Replacing VLM responses with empty or constant strings yields no predictions and zero F1 for all three selected harnesses. Substituting \vlm with Kimi-VL, without repeating the search and keeping the harness fixed yields $0.154 \pm 0.071$ (\cref{tab:suppl:vlm_control}).

\paragraph{Results across VLMs and orchestrators (\cref{tab:suppl:vlms,tab:suppl:grid}).}
\method generalises across VLM--orchestrator combinations. With \orch as orchestrator, it outperforms \spotlight's zero-shot and MA baselines for every VLM (\cref{tab:suppl:vlms}). Performance also depends on the orchestrator: \orch achieves the highest test F1 with all four VLMs (\cref{tab:suppl:grid}). GLM~5.3~Flash performs comparably with Kimi-VL (0.215 versus 0.219), but reaches 0.157--0.164 with the remaining VLMs. MiniMax~M3.1~Flash achieves 0.134--0.178, peaking with Kimi-VL. Overall, all twelve combinations outperform baselines on the test split, and ten on the OOD split.

\section{Conclusion}
\label{sec:conclusion}

We formulate VLM harness design for generated-video error localisation as a search over executable programs, optimising F1 to penalise unmatched predictions overlooked by the benchmark's metric. \method combines a search memory, independent lineages over random splits, and constructive merging, with candidates ranked on a hidden validation set. Under a common judging protocol, the searched harness achieves the highest F1 and precision compared with other baselines -- \spotlight benchmark strategies, fine-tuning, prompt optimisation, frontier models, and adaptations of recent agentic harnesses. These gains extend to an unseen generator and different VLMs, while ablations highlight the contributions of the individual design choices. 
We further evaluate cross-benchmark transfer and in-domain optimisation on IPV-Bench.

\paragraph{Limitations and Future Work}
\textit{(i) Labelled data:} Search requires a labelled training pool, as the orchestrator accesses training annotations.
\textit{(ii) Judge dependence:} All results rely on \judge and the benchmark's matching rule, inheriting its biases.
\textit{(iii) Orchestrator sensitivity:}
\orch achieves the highest test F1 with all four VLMs,
outperforming GLM~5.3~Flash and MiniMax~M3.1~Flash in most
settings (\cref{tab:suppl:grid}). Extending \method beyond error localisation to search for general-purpose harnesses across a broader range of tasks is a promising direction.

\section*{LLM Usage Statement}
In this work, we used generative AI tools to assist in implementing the proposed method and the baselines, in running and monitoring experiments, and in analysing experimental results. We also used generative AI tools to draft portions of the manuscript, to assist in producing tables and result visualisations, and to improve readability through grammar, spelling, and word-choice edits. We have not used generative AI tools to develop the research ideas or to formulate the conclusions. All AI-assisted code, analyses, and manuscript drafts were reviewed by the authors, and results were checked against the stored experimental outputs. The research ideas, the choice of experiments, and the conclusions are the authors' own. We take responsibility for the final content of this work, including text, claims, code, and artifacts produced with the aid of generative AI.
{
    \small
    \bibliographystyle{ieeenat_fullname}
    \bibliography{main}

@inproceedings{huang2024vbench,
  title={Vbench: Comprehensive benchmark suite for video generative models},
  author={Huang, Ziqi and He, Yinan and Yu, Jiashuo and Zhang, Fan and Si, Chenyang and Jiang, Yuming and Zhang, Yuanhan and Wu, Tianxing and Jin, Qingyang and Chanpaisit, Nattapol and others},
  booktitle={2024 IEEE/CVF Conference on Computer Vision and Pattern Recognition (CVPR)},
  pages={21807--21818},
  year={2024},
  organization={IEEE}
}

@article{zheng2503vbench,
  title={Vbench-2.0: Advancing video generation benchmark suite for intrinsic faithfulness, 2025},
  author={Zheng, Dian and Huang, Ziqi and Liu, Hongbo and Zou, Kai and He, Yinan and Zhang, Fan and Gu, Lulu and Zhang, Yuanhan and He, Jingwen and Zheng, Wei-Shi and others},
  journal={URL https://arxiv. org/abs/2503.21755}
}

@inproceedings{liu2024evalcrafter,
  title={Evalcrafter: Benchmarking and evaluating large video generation models},
  author={Liu, Yaofang and Cun, Xiaodong and Liu, Xuebo and Wang, Xintao and Zhang, Yong and Chen, Haoxin and Liu, Yang and Zeng, Tieyong and Chan, Raymond and Shan, Ying},
  booktitle={2024 IEEE/CVF Conference on Computer Vision and Pattern Recognition (CVPR)},
  pages={22139--22149},
  year={2024},
  organization={IEEE}
}

@inproceedings{chinchure2026spotlight,
  title={Spotlight: Identifying and localizing video generation errors using vlms},
  author={Chinchure, Aditya and Ravi, Sahithya and Shukla, Pushkar and Shwartz, Vered and Sigal, Leonid},
  booktitle={European Conference on Computer Vision},
  pages={1--17},
  year={2026},
  organization={Springer}
}

@inproceedings{zhang2025evaluation,
  title={Evaluation agent: Efficient and promptable evaluation framework for visual generative models},
  author={Zhang, Fan and Tian, Shulin and Huang, Ziqi and Qiao, Yu and Liu, Ziwei},
  booktitle={Proceedings of the 63rd Annual Meeting of the Association for Computational Linguistics (Volume 1: Long Papers)},
  pages={7561--7582},
  year={2025}
}

@article{tian2026open,
  title={Open Evaluation Agent: Efficient and Promptable Evaluation of Visual Generative Models},
  author={Tian, Shulin and Huang, Ziqi and Zhang, Fan and Zhu, Hongyuan and Qiao, Yu and Liu, Ziwei},
  journal={arXiv preprint arXiv:2608.09666},
  year={2026}
}

@misc{videoargus,
  title         = {VideoArgus: Agentic Rubric-Grounded Unified Evaluation for Video Generation and Editing},
  author        = {Ziyun Zeng and Zixuan Wang and Yongsheng Yu and Hang Hua and Jiebo Luo},
  year          = {2026},
  eprint        = {2608.05485},
  archivePrefix = {arXiv},
  primaryClass  = {cs.CV},
  howpublished  = {arXiv:2608.05485},
}

@article{song2026vqqa,
  title={Vqqa: An agentic approach for video evaluation and quality improvement},
  author={Song, Yiwen and Pfister, Tomas and Song, Yale},
  journal={arXiv preprint arXiv:2603.12310},
  year={2026}
}

@article{tang2026rewardverse,
  title={RewardVerse: Rubric-Guided Policy Optimization for Video Reward Modeling},
  author={Tang, Zhenchen and Li, Yang and Yang, Songlin and Peng, Bo and Zhao, Xiaotong and Li, Shuai and Fan, Haotian and Zhao, Alan and Dong, Jing},
  journal={arXiv preprint arXiv:2609.22947},
  year={2026}
}

@article{wei2026videoweaver,
  title={VideoWeaver: Evaluating and Evolving Skills for Agentic Long Video Generation},
  author={Wei, Jianhui and Tan, Jie and Zhu, Hengchuan and Zhang, Xiaotian and Zhang, Yan and Chen, Ziyi and Zhang, Daoan and Xu, Wei and Liu, Zuozhu},
  journal={arXiv preprint arXiv:2606.08091},
  year={2026}
}

@inproceedings{long2026vista,
  title={Vista: A test-time self-improving video generation agent},
  author={Long, Do Xuan and Wan, Xingchen and Nakhost, Hootan and Lee, Chen-Yu and Pfister, Tomas and Arik, Sercan {\"O}},
  booktitle={Proceedings of the IEEE/CVF conference on computer vision and pattern recognition},
  pages={6021--6032},
  year={2026}
}

@article{xu2026don,
  title={Don't Read the Log: Execution Traces Contaminate Verifiers in Video-Generation Agents},
  author={Xu, Jian},
  journal={arXiv preprint arXiv:2609.28564},
  year={2026}
}

@article{cao2026each,
  title={Each Judge Its Own Yardstick: Discovering Per-VLM Taxonomies for Physical Video Evaluation},
  author={Cao, Yu and Liu, Ziquan and Zhang, Zhensong and Deng, Jiankang and Gong, Shaogang and Song, Jifei},
  journal={arXiv preprint arXiv:2606.22918},
  year={2026}
}

@inproceedings{agrawal2026gepa,
  title={Gepa: Reflective prompt evolution can outperform reinforcement learning},
  author={Agrawal, Lakshya A and Tan, Shangyin and Soylu, Dilara and Ziems, Noah and Khare, Rishi and Opsahl-Ong, Krista and Singhvi, Arnav and Shandilya, Herumb and Ryan, Michael J and Jiang, Meng and others},
  booktitle={International Conference on Learning Representations},
  volume={2026},
  pages={8479--8565},
  year={2026}
}

@article{yuksekgonul2024textgrad,
  title={Textgrad: Automatic" differentiation" via text},
  author={Yuksekgonul, Mert and Bianchi, Federico and Boen, Joseph and Liu, Sheng and Huang, Zhi and Guestrin, Carlos and Zou, James},
  journal={arXiv preprint arXiv:2406.07496},
  year={2024}
}

@inproceedings{yang2024large,
  title={Large language models as optimizers},
  author={Yang, Chengrun and Wang, Xuezhi and Lu, Yifeng and Liu, Hanxiao and Le, Quoc V and Zhou, Denny and Chen, Xinyun},
  booktitle={International Conference on Learning Representations},
  volume={2024},
  pages={12028--12068},
  year={2024}
}

@article{shinn2023reflexion,
  title={Reflexion: Language agents with verbal reinforcement learning},
  author={Shinn, Noah and Cassano, Federico and Gopinath, Ashwin and Narasimhan, Karthik and Yao, Shunyu},
  journal={Advances in neural information processing systems},
  volume={36},
  pages={8634--8652},
  year={2023}
}

@article{madaan2023self,
  title={Self-refine: Iterative refinement with self-feedback},
  author={Madaan, Aman and Tandon, Niket and Gupta, Prakhar and Hallinan, Skyler and Gao, Luyu and Wiegreffe, Sarah and Alon, Uri and Dziri, Nouha and Prabhumoye, Shrimai and Yang, Yiming and others},
  journal={Advances in neural information processing systems},
  volume={36},
  pages={46534--46594},
  year={2023}
}

@article{novikov2025alphaevolve,
  title={Alphaevolve: A coding agent for scientific and algorithmic discovery},
  author={Novikov, Alexander and V{\~u}, Ng{\^a}n and Eisenberger, Marvin and Dupont, Emilien and Huang, Po-Sen and Wagner, Adam Zsolt and Shirobokov, Sergey and Kozlovskii, Borislav and Ruiz, Francisco JR and Mehrabian, Abbas and others},
  journal={arXiv preprint arXiv:2506.13131},
  year={2025}
}

@article{ye2026videoresearcher,
  title={VideoResearcher: Self-Improving Tool Design for Long-Video Understanding},
  author={Ye, Dingqiang and Zhao, Dongdi and Wang, Kaishen and Hu, Qingqiao and Sun, Jingchen and Liang, Yijun and Jia, Yuqi and Huang, Yiqiao and Tian, Yunjie and Zhang, Jiaxing and others},
  journal={arXiv preprint arXiv:2609.19664},
  year={2026}
}

@article{bai2025impossible,
  title={Impossible videos},
  author={Bai, Zechen and Ci, Hai and Shou, Mike Zheng},
  journal={arXiv preprint arXiv:2503.14378},
  year={2025}
}

@article{hu2021lora,
  title={Lora: Low-rank adaptation of large language models},
  author={Hu, Edward J and Shen, Yelong and Wallis, Phillip and Allen-Zhu, Zeyuan and Li, Yuanzhi and Wang, Shean and Wang, Lu and Chen, Weizhu},
  journal={arXiv preprint arXiv:2106.09685},
  year={2021}
}

@inproceedings{opsahl2024optimizing,
  title={Optimizing instructions and demonstrations for multi-stage language model programs},
  author={Opsahl-Ong, Krista and Ryan, Michael J and Purtell, Josh and Broman, David and Potts, Christopher and Zaharia, Matei and Khattab, Omar},
  booktitle={Proceedings of the 2024 Conference on Empirical Methods in Natural Language Processing},
  pages={9340--9366},
  year={2024}
}

@article{xu2026harness,
  title   = {Harness$^2$: Recursive Agent Harnessing for an Open World},
  author  = {Xu, Ruiyao and Chen, Yanfei and CuiZhu, Zhongying and
             Dalvi Mishra, Bhavana and Ming, Yifei and Yu, Han and
             Han, Rujun and Lee, Chen-Yu and Pfister, Tomas},
  journal = {arXiv preprint arXiv:2610.14224},
  year    = {2026},
  url     = {https://arxiv.org/abs/2610.14224}
}

@inproceedings{hu2025automated,
  title={Automated design of agentic systems},
  author={Hu, Shengran and Lu, Cong and Clune, Jeff},
  booktitle={International Conference on Learning Representations},
  volume={2025},
  pages={21344--21377},
  year={2025}
}

@inproceedings{zhang2025aflow,
  title={Aflow: Automating agentic workflow generation},
  author={Zhang, Jiayi and Xiang, Jinyu and Yu, Zhaoyang and Teng, Fengwei and Chen, Xionghui and Chen, Jiaqi and Zhuge, Mingchen and Cheng, Xin and Hong, Sirui and Wang, Jinlin and others},
  booktitle={International Conference on Learning Representations},
  volume={2025},
  pages={34040--34077},
  year={2025}
}

@inproceedings{lange2026shinkaevolve,
  title={Shinkaevolve: Towards open-ended and sample-efficient program evolution},
  author={Lange, Robert and Imajuku, Yuki and Cetin, Edoardo},
  booktitle={International Conference on Learning Representations},
  volume={2026},
  pages={74026--74078},
  year={2026}
}

@software{openevolve,
  title = {OpenEvolve: Open-source implementation of AlphaEvolve},
  author = {Asankhaya Sharma},
  year = {2025},
  publisher = {GitHub},
  url = {https://github.com/codelion/openevolve}
}
}

\clearpage
\nocite{lange2026shinkaevolve}
\setcounter{page}{1}
\maketitlesupplementary
\setcounter{table}{0}\renewcommand{\thetable}{S\arabic{table}}
\setcounter{figure}{0}\renewcommand{\thefigure}{S\arabic{figure}}
\renewcommand{\theHtable}{S\arabic{table}}\renewcommand{\theHfigure}{S\arabic{figure}}

\section{Extended results}
\label{sec:suppl:results_full}
\Cref{tab:suppl:results_full} reports the scalar score, $\Mpct$, precision, F1, and prediction count for every method in \cref{tab:results} on the test and LTX-2 OOD splits, along with additional results using Kimi-VL as a second backbone. \Cref{tab:suppl:sd} reports standard deviations across three runs for these metrics and for test F1 in the search design ablation (\cref{tab:ablation}). \Cref{tab:suppl:categories,tab:suppl:categories_baselines} further report, by \spotlight error category, recall for the harness in \cref{fig:overview} and F1 for \method and the baselines, respectively.

\begin{table*}[t]
\centering
\small
\setlength{\tabcolsep}{1.9pt}
\resizebox{\textwidth}{!}{%
\begin{tabular}{l*{5}{c>{\columncolor{oodshade}}c>{\columncolor{oodshade}}c}}
\toprule
 & \multicolumn{3}{c}{Scalar\,$\uparrow$} & \multicolumn{3}{c}{\Mpct\,$\uparrow$} & \multicolumn{3}{c}{Prec\,$\uparrow$} & \multicolumn{3}{c}{F1\,$\uparrow$} & \multicolumn{3}{c}{\#pred} \\
\cmidrule(lr){2-4}\cmidrule(lr){5-7}\cmidrule(lr){8-10}\cmidrule(lr){11-13}\cmidrule(lr){14-16}
Method & Test & OOD & OOD-40 & Test & OOD & OOD-40 & Test & OOD & OOD-40 & Test & OOD & OOD-40 & Test & OOD & OOD-40 \\
\midrule
\multicolumn{16}{l}{\emph{Qwen3-VL-8B}} \\
ZS (A6) & .172\sd{.030} & .241\sd{.003} & .222\sd{.013} & .130\sd{.027} & .165\sd{.004} & .169\sd{.015} & .133\sd{.019} & .144\sd{.002} & .125\sd{.013} & .119\sd{.020} & .132\sd{.001} & .130\sd{.008} & 249 & 737 & 166 \\
MA (A6) & .436\sd{.012} & .457\sd{.005} & .468\sd{.014} & .383\sd{.012} & .375\sd{.007} & .405\sd{.017} & .114\sd{.006} & .115\sd{.001} & .099\sd{.004} & .164\sd{.008} & .163\sd{.001} & .149\sd{.005} & 702 & 1953 & 417 \\
\midrule
\multicolumn{16}{l}{\emph{Qwen3-VL-30B}} \\
ZS (A7) & .245\sd{.017} & .304\sd{.004} & .280\sd{.018} & .192\sd{.013} & .229\sd{.007} & .215\sd{.016} & .093\sd{.014} & .119\sd{.003} & .087\sd{.008} & .107\sd{.012} & .135\sd{.003} & .117\sd{.008} & 655 & 1623 & 361 \\
\midrule
\multicolumn{16}{l}{\emph{Gemini-2.5-Pro}} \\
ZS (A8) & .360\sd{.021} & .364\sd{.007} & .339\sd{.035} & .294\sd{.023} & .279\sd{.015} & .262\sd{.019} & .125\sd{.009} & .123\sd{.007} & .122\sd{.017} & .167\sd{.010} & .160\sd{.008} & .158\sd{.019} & 461 & 1247 & 252 \\
\midrule
\multicolumn{16}{l}{\emph{DeepSeek-V4.1}} \\
ZS (A8b) & .377\sd{.034} & .369\sd{.032} & .366\sd{.053} & .299\sd{.029} & .277\sd{.027} & .290\sd{.055} & .140\sd{.028} & .125\sd{.007} & .115\sd{.012} & .175\sd{.029} & .157\sd{.011} & .158\sd{.018} & 454 & 1308 & 252 \\
\midrule
\multicolumn{16}{l}{\emph{GLM-5.3-Flash}} \\
ZS (A8c) & .412\sd{.018} & .393\sd{.009} & .342\sd{.031} & .328\sd{.031} & .295\sd{.005} & .257\sd{.026} & .157\sd{.005} & .157\sd{.009} & .136\sd{.017} & .200\sd{.008} & .191\sd{.008} & .166\sd{.019} & 410 & 1077 & 215 \\
\midrule
\multicolumn{16}{l}{\emph{MiniMax-M3.1-Flash}} \\
ZS (A8d) & .454\sd{.050} & .436\sd{.012} & .394\sd{.069} & .381\sd{.042} & .350\sd{.014} & .335\sd{.073} & .135\sd{.011} & .141\sd{.008} & .123\sd{.025} & .187\sd{.018} & .185\sd{.008} & .170\sd{.033} & 571 & 1402 & 282 \\
\midrule
\multicolumn{16}{l}{\emph{\vlm}} \\
ZS (A1) & .207\sd{.011} & .226\sd{.008} & .199\sd{.016} & .171\sd{.010} & .153\sd{.006} & .137\sd{.020} & .135\sd{.005} & .132\sd{.006} & .104\sd{.013} & .141\sd{.007} & .125\sd{.005} & .110\sd{.014} & 237 & 667 & 119 \\
MA (A2) & .434\sd{.017} & .457\sd{.016} & .469\sd{.016} & .373\sd{.008} & .384\sd{.017} & .409\sd{.025} & .090\sd{.011} & .105\sd{.005} & .105\sd{.004} & .130\sd{.009} & .139\sd{.006} & .155\sd{.004} & 1104 & 2858 & 618 \\
Seq (A2) & .236\sd{.005} & .207\sd{.001} & .226\sd{.021} & .175\sd{.003} & .144\sd{.002} & .158\sd{.013} & .160\sd{.003} & .129\sd{.002} & .119\sd{.012} & .155\sd{.002} & .122\sd{.002} & .125\sd{.013} & 310 & 743 & 141 \\
SW (A2) & .399\sd{.010} & .416\sd{.004} & .380\sd{.014} & .324\sd{.014} & .322\sd{.005} & .302\sd{.014} & .090\sd{.003} & .092\sd{.004} & .075\sd{.006} & .132\sd{.003} & .133\sd{.005} & .113\sd{.008} & 733 & 2168 & 473 \\
CoT (A2) & .271\sd{.025} & .242\sd{.011} & .292\sd{.046} & .218\sd{.022} & .175\sd{.008} & .216\sd{.044} & .171\sd{.013} & .144\sd{.006} & .171\sd{.022} & .173\sd{.016} & .140\sd{.003} & .172\sd{.026} & 266 & 678 & 140 \\
\addlinespace[2pt]
LoRA (A3) & .226\sd{.047} & .262\sd{.018} & .236\sd{.015} & .175\sd{.032} & .177\sd{.009} & .163\sd{.014} & .131\sd{.036} & .154\sd{.016} & .138\sd{.020} & .136\sd{.035} & .148\sd{.010} & .140\sd{.015} & 267 & 675 & 126 \\
LoRA-F1 (A3) & .254\sd{.049} & .274\sd{.021} & .230\sd{.010} & .191\sd{.040} & .186\sd{.014} & .156\sd{.013} & .155\sd{.021} & .169\sd{.019} & .131\sd{.020} & .154\sd{.025} & .159\sd{.017} & .134\sd{.015} & 246 & 631 & 121 \\
MIPROv2 (A4) & .214\sd{.013} & .247\sd{.010} & .262\sd{.059} & .172\sd{.011} & .174\sd{.012} & .198\sd{.042} & .105\sd{.018} & .122\sd{.021} & .125\sd{.044} & .120\sd{.013} & .122\sd{.009} & .133\sd{.029} & 340 & 970 & 227 \\
MIPROv2-F1 (A4) & .227\sd{.005} & .228\sd{.050} & .242\sd{.077} & .175\sd{.002} & .160\sd{.042} & .182\sd{.063} & .154\sd{.021} & .147\sd{.020} & .147\sd{.050} & .141\sd{.008} & .130\sd{.026} & .141\sd{.045} & 286 & 741 & 152 \\
\addlinespace[2pt]
JudgeFit (A11) & .529\sd{.037} & .506\sd{.018} & .517\sd{.040} & .464\sd{.041} & .431\sd{.018} & .470\sd{.039} & .074\sd{.003} & .072\sd{.005} & .083\sd{.019} & .121\sd{.006} & .111\sd{.006} & .121\sd{.019} & 1464 & 4100 & 889 \\
JudgeFit-F1 (A11) & .440\sd{.051} & .457\sd{.004} & .476\sd{.030} & .373\sd{.052} & .380\sd{.011} & .432\sd{.033} & .090\sd{.008} & .101\sd{.021} & .105\sd{.028} & .134\sd{.007} & .141\sd{.020} & .153\sd{.028} & 940 & 2607 & 550 \\
VideoArgus (A12) & .146\sd{.010} & .134\sd{.010} & .132\sd{.020} & .114\sd{.005} & .092\sd{.007} & .096\sd{.015} & .089\sd{.009} & .073\sd{.011} & .057\sd{.023} & .087\sd{.007} & .069\sd{.008} & .065\sd{.022} & 199 & 470 & 102 \\
Harness$^2$ (A13) & .120\sd{.051} & .121\sd{.037} & .114\sd{.033} & .092\sd{.048} & .078\sd{.029} & .068\sd{.028} & .109\sd{.035} & .103\sd{.043} & .111\sd{.042} & .087\sd{.038} & .074\sd{.026} & .073\sd{.027} & 156 & 435 & 86 \\
\cmidrule{1-16}
\method (Deepseek-orch) & .398\sd{.022} & .368\sd{.050} & .375\sd{.084} & .327\sd{.024} & .284\sd{.051} & .305\sd{.079} & .203\sd{.028} & .197\sd{.017} & .190\sd{.006} & .234\sd{.012} & .213\sd{.008} & .217\sd{.028} & 309 & 792 & 161 \\
\method-Full (Deepseek-orch) & .397\sd{.048} & .394\sd{.069} & .399\sd{.020} & .336\sd{.052} & .304\sd{.060} & .327\sd{.025} & .221\sd{.012} & .235\sd{.017} & .228\sd{.032} & .241\sd{.016} & .239\sd{.015} & .244\sd{.011} & 283 & 721 & 149 \\
\method (Qwen-orch) & .503\sd{.134} & .557\sd{.140} & .628\sd{.164} & .462\sd{.144} & .478\sd{.145} & .558\sd{.161} & .259\sd{.051} & .249\sd{.041} & .246\sd{.049} & .284\sd{.049} & .290\sd{.042} & .314\sd{.063} & 433 & 1318 & 272 \\
\method-Full (Qwen-orch) & .546\sd{.081} & .587\sd{.128} & .356\sd{.113} & .504\sd{.116} & .521\sd{.153} & .291\sd{.104} & \textbf{.298}\sd{.065} & \textbf{.274}\sd{.037} & .292\sd{.021} & \textbf{.340}\sd{.008} & \textbf{.321}\sd{.026} & .278\sd{.080} & 358 & --- & 108 \\
HSM$^*$ & \textbf{.824} & \textbf{.842} & \textbf{.874} & \textbf{.840} & \textbf{.857} & \textbf{.904} & .002 & .002 & .002 & .003 & .004 & .004 & 108062 & 275127 & 54532 \\
\midrule
\multicolumn{16}{l}{\emph{Kimi-VL-A3B}} \\
ZS (A1) & .178\sd{.006} & .170\sd{.009} & .194\sd{.017} & .145\sd{.011} & .130\sd{.008} & .176\sd{.014} & .066\sd{.005} & .055\sd{.005} & .062\sd{.010} & .066\sd{.004} & .063\sd{.005} & .081\sd{.011} & 506 & 1707 & 347 \\
MA (A2) & .465\sd{.007} & .542\sd{.004} & .584\sd{.007} & .420\sd{.012} & .480\sd{.005} & .544\sd{.013} & .054\sd{.003} & .060\sd{.003} & .067\sd{.002} & .088\sd{.003} & .097\sd{.004} & .107\sd{.005} & 1978 & 6746 & 1443 \\
Seq (A2) & .203\sd{.010} & .200\sd{.014} & .215\sd{.031} & .149\sd{.007} & .147\sd{.012} & .171\sd{.033} & .172\sd{.009} & .160\sd{.014} & .156\sd{.016} & .144\sd{.005} & .129\sd{.012} & .136\sd{.021} & 272 & 600 & 109 \\
SW (A2) & .336\sd{.008} & .352\sd{.008} & .291\sd{.007} & .281\sd{.005} & .271\sd{.009} & .206\sd{.012} & .060\sd{.001} & .047\sd{.001} & .040\sd{.006} & .092\sd{.002} & .072\sd{.001} & .064\sd{.008} & 1015 & 3800 & 788 \\
CoT (A2) & .251\sd{.015} & .217\sd{.012} & .144\sd{.037} & .222\sd{.019} & .168\sd{.006} & .114\sd{.025} & .059\sd{.007} & .047\sd{.000} & .025\sd{.005} & .084\sd{.007} & .064\sd{.003} & .038\sd{.008} & 724 & 1743 & 360 \\
\addlinespace[2pt]
LoRA (A3) & .118\sd{.017} & .163\sd{.032} & .135\sd{.029} & .095\sd{.013} & .119\sd{.029} & .096\sd{.026} & .070\sd{.006} & .123\sd{.008} & .096\sd{.006} & .070\sd{.005} & .102\sd{.011} & .079\sd{.006} & 250 & 576 & 127 \\
LoRA-F1 (A3) & .103\sd{.043} & .145\sd{.050} & .119\sd{.045} & .080\sd{.030} & .101\sd{.042} & .087\sd{.047} & .071\sd{.022} & .116\sd{.025} & .080\sd{.024} & .067\sd{.022} & .091\sd{.030} & .070\sd{.031} & 204 & 505 & 97 \\
MIPROv2 (A4) & .261\sd{.035} & .270\sd{.041} & .217\sd{.008} & .230\sd{.029} & .215\sd{.037} & .176\sd{.013} & .060\sd{.002} & .049\sd{.006} & .039\sd{.001} & .090\sd{.006} & .074\sd{.009} & .061\sd{.001} & 742 & 2337 & 463 \\
MIPROv2-F1 (A4) & .219\sd{.033} & .193\sd{.036} & .221\sd{.063} & .179\sd{.034} & .145\sd{.032} & .178\sd{.055} & .065\sd{.011} & .068\sd{.007} & .080\sd{.015} & .082\sd{.008} & .070\sd{.007} & .087\sd{.014} & 641 & 1751 & 372 \\
\addlinespace[2pt]
JudgeFit (A11) & \textbf{.543}\sd{.032} & \textbf{.588}\sd{.024} & \textbf{.585}\sd{.034} & \textbf{.497}\sd{.041} & \textbf{.534}\sd{.032} & \textbf{.549}\sd{.034} & .049\sd{.005} & .038\sd{.001} & .039\sd{.005} & .083\sd{.008} & .068\sd{.001} & .068\sd{.007} & 2709 & 9863 & 2028 \\
JudgeFit-F1 (A11) & .454\sd{.092} & .502\sd{.063} & .470\sd{.075} & .397\sd{.104} & .438\sd{.067} & .421\sd{.084} & .079\sd{.010} & .068\sd{.003} & .064\sd{.011} & .116\sd{.015} & .107\sd{.004} & .100\sd{.017} & 1353 & 4812 & 1020 \\
VideoArgus (A12) & .053\sd{.010} & .093\sd{.017} & .089\sd{.050} & .039\sd{.007} & .071\sd{.019} & .063\sd{.046} & .025\sd{.005} & .044\sd{.009} & .047\sd{.038} & .021\sd{.003} & .037\sd{.009} & .038\sd{.030} & 150 & 810 & 177 \\
Harness$^2$ (A13) & .065\sd{.049} & .061\sd{.060} & .061\sd{.075} & .049\sd{.047} & .046\sd{.046} & .051\sd{.059} & .044\sd{.014} & .052\sd{.044} & .054\sd{.066} & .036\sd{.018} & .038\sd{.034} & .041\sd{.048} & 158 & 376 & 79 \\
\cmidrule{1-16}
\method (Ours) & .309\sd{.045} & .317\sd{.029} & .297\sd{.018} & .248\sd{.048} & .234\sd{.033} & .213\sd{.023} & \textbf{.242}\sd{.052} & \textbf{.230}\sd{.050} & \textbf{.207}\sd{.084} & \textbf{.219}\sd{.028} & \textbf{.202}\sd{.008} & \textbf{.183}\sd{.028} & 209 & 572 & 124 \\

\bottomrule
\end{tabular}}
\caption{\textbf{Full results on \spotlight.} Evaluation uses the 80-clip test split, the 200 OOD LTX-2 clips
(\cref{sec:results:setup}) and OOD-40, the 40 OOD clips whose prompts are the test prompts. Rows are grouped by video
model and compare baselines and \method (\cref{sec:results:baselines}); \#pred is the number of predictions on the split.
Values with $\pm$ are the mean and standard deviation over three runs (\#pred is the mean); rows without $\pm$ are single
runs. Bold indicates the best value per column, separately for the Kimi-VL block and the rows above it.}
\label{tab:suppl:results_full}
\end{table*}

\begin{table}[t]
\centering
\footnotesize
\setlength{\tabcolsep}{2pt}
\resizebox{\linewidth}{!}{%
\begin{tabular}{l*{5}{c>{\columncolor{oodshade}}c}}
\toprule
 & \multicolumn{2}{c}{Scalar\,$\uparrow$} & \multicolumn{2}{c}{\Mpct\,$\uparrow$} & \multicolumn{2}{c}{Prec\,$\uparrow$} & \multicolumn{2}{c}{F1\,$\uparrow$} & \multicolumn{2}{c}{pred/clip} \\
\cmidrule(lr){2-3}\cmidrule(lr){4-5}\cmidrule(lr){6-7}\cmidrule(lr){8-9}\cmidrule(lr){10-11}
Method (\vlm) & Test & OOD & Test & OOD & Test & OOD & Test & OOD & Test & OOD \\
\midrule
ZS (A1) & .207 & .226 & .171 & .153 & .135 & .132 & .141 & .125 & 3.0 & 3.3 \\
MA (A2) & .434 & .457 & .373 & .384 & .090 & .105 & .130 & .139 & 13.8 & 14.3 \\
Seq (A2) & .236 & .207 & .175 & .144 & .160 & .129 & .155 & .122 & 3.9 & 3.7 \\
SW (A2) & .399 & .416 & .324 & .322 & .090 & .092 & .132 & .133 & 9.2 & 10.8 \\
CoT (A2) & .271 & .242 & .218 & .175 & .171 & .144 & .173 & .140 & 3.3 & 3.4 \\
\addlinespace[2pt]
LoRA (A3) & .226 & .262 & .175 & .177 & .131 & .154 & .136 & .148 & 3.3 & 3.4 \\
LoRA-F1 (A3) & .254 & .274 & .191 & .186 & .155 & .169 & .154 & .159 & 3.1 & 3.2 \\
MIPROv2 (A4) & .214 & .247 & .172 & .174 & .105 & .122 & .120 & .122 & 4.2 & 4.8 \\
MIPROv2-F1 (A4) & .227 & .228 & .175 & .160 & .154 & .147 & .141 & .130 & 3.6 & 3.7 \\
\addlinespace[2pt]
JudgeFit (A11) & .529 & .506 & .464 & .431 & .074 & .072 & .121 & .111 & 18.3 & 20.5 \\
JudgeFit-F1 (A11) & .440 & .457 & .373 & .380 & .090 & .101 & .134 & .141 & 11.8 & 13.0 \\
VideoArgus (A12) & .146 & .134 & .114 & .092 & .089 & .073 & .087 & .069 & 2.5 & 2.4 \\
Harness$^2$ (A13) & .120 & .121 & .092 & .078 & .109 & .103 & .087 & .074 & 1.9 & 2.2 \\
\cmidrule{1-11}
\method (Deepseek-orch) & .398 & .368 & .327 & .284 & .203 & .197 & .234 & .213 & 3.9 & 4.0 \\
\method-Full (Deepseek-orch) & .397 & .394 & .336 & .304 & .221 & .235 & .241 & .239 & 3.5 & 3.6 \\
\method (Qwen-orch) & .503 & .557 & .462 & .478 & .259 & .249 & .284 & .290 & 5.4 & 6.6 \\
\method-Full (Qwen-orch) & .546 & .587 & .504 & .521 & \textbf{.298} & \textbf{.274} & \textbf{.340} & \textbf{.321} & 4.5 & 5.7 \\
HSM$^*$ & \textbf{.824} & \textbf{.842} & \textbf{.840} & \textbf{.857} & .002 & .002 & .003 & .004 & 1351 & 1376 \\
\bottomrule
\end{tabular}}
\caption{\textbf{Full metrics with \vlm.} Evaluation uses the 80-clip test split and the 200 OOD LTX-2 clips (\cref{sec:results:setup}); pred/clip is the mean number of predictions per clip. Values are averaged over three runs (standard deviations in \cref{tab:suppl:sd}), except HSM$^*$, which are single runs. HSM$^*$ changes only the objective to $(M+\Mpct)/2$, which results in over-prediction ($>1{,}000$ predictions per clip). Bold indicates the best value per column.}
\label{tab:results_vlm}
\end{table}

\begin{table*}[t]
\centering
\scriptsize
\begin{subtable}{\textwidth}
\centering
\setlength{\tabcolsep}{1.9pt}
\resizebox{\textwidth}{!}{%
\begin{tabular}{l*{5}{c>{\columncolor{oodshade}}c>{\columncolor{oodshade}}c}}
\toprule
 & \multicolumn{3}{c}{Scalar} & \multicolumn{3}{c}{\Mpct} & \multicolumn{3}{c}{Prec} & \multicolumn{3}{c}{F1} & \multicolumn{3}{c}{\#pred} \\
\cmidrule(lr){2-4}\cmidrule(lr){5-7}\cmidrule(lr){8-10}\cmidrule(lr){11-13}\cmidrule(lr){14-16}
Method & Test & OOD & OOD-40 & Test & OOD & OOD-40 & Test & OOD & OOD-40 & Test & OOD & OOD-40 & Test & OOD & OOD-40 \\
\midrule
\multicolumn{16}{l}{\emph{Baselines with \vlm}} \\
ZS (A1) & .011 & .008 & .016 & .010 & .006 & .020 & .005 & .006 & .013 & .007 & .005 & .014 & 7 & 7 & 2 \\
MA (A2) & .017 & .016 & .016 & .008 & .017 & .025 & .011 & .005 & .004 & .009 & .006 & .004 & 33 & 46 & 20 \\
Seq (A2) & .005 & .001 & .021 & .003 & .002 & .013 & .003 & .002 & .012 & .002 & .002 & .013 & 25 & 28 & 4 \\
SW (A2) & .010 & .004 & .014 & .014 & .005 & .014 & .003 & .004 & .006 & .003 & .005 & .008 & 21 & 26 & 16 \\
CoT (A2) & .025 & .011 & .046 & .022 & .008 & .044 & .013 & .006 & .022 & .016 & .003 & .026 & 18 & 27 & 3 \\
LoRA (A3) & .047 & .018 & .015 & .032 & .009 & .014 & .036 & .016 & .020 & .035 & .010 & .015 & 39 & 76 & 10 \\
LoRA-F1 (A3) & .049 & .021 & .010 & .040 & .014 & .013 & .021 & .019 & .020 & .025 & .017 & .015 & 8 & 4 & 3 \\
MIPROv2 (A4) & .013 & .010 & .059 & .011 & .012 & .042 & .018 & .021 & .044 & .013 & .009 & .029 & 98 & 216 & 53 \\
MIPROv2-F1 (A4) & .005 & .050 & .077 & .002 & .042 & .063 & .021 & .020 & .050 & .008 & .026 & .045 & 78 & 131 & 33 \\
JudgeFit (A11) & .037 & .018 & .040 & .041 & .018 & .039 & .003 & .005 & .019 & .006 & .006 & .019 & 93 & 202 & 67 \\
JudgeFit-F1 (A11) & .051 & .004 & .030 & .052 & .011 & .033 & .008 & .021 & .028 & .007 & .020 & .028 & 178 & 418 & 101 \\
VideoArgus (A12) & .010 & .010 & .020 & .005 & .007 & .015 & .009 & .011 & .023 & .007 & .008 & .022 & 16 & 16 & 18 \\
Harness$^2$ (A13) & .051 & .037 & .033 & .048 & .029 & .028 & .035 & .043 & .042 & .038 & .026 & .027 & 108 & 304 & 62 \\
\midrule
\multicolumn{16}{l}{\emph{Prompts with other VLMs}} \\
Qwen3-VL-8B, ZS (A6) & .030 & .003 & .013 & .027 & .004 & .015 & .019 & .002 & .013 & .020 & .001 & .008 & 14 & 25 & 27 \\
Qwen3-VL-8B, MA (A6) & .012 & .005 & .014 & .012 & .007 & .017 & .006 & .001 & .004 & .008 & .001 & .005 & 19 & 9 & 9 \\
Qwen3-VL-30B, ZS (A7) & .017 & .004 & .018 & .013 & .007 & .016 & .014 & .003 & .008 & .012 & .003 & .008 & 42 & 52 & 53 \\
Gemini-2.5-Pro, ZS (A8) & .021 & .007 & .035 & .023 & .015 & .019 & .009 & .007 & .017 & .010 & .008 & .019 & 25 & 21 & 2 \\
DeepSeek-V4.1, ZS (A8b) & .034 & .032 & .053 & .029 & .027 & .055 & .028 & .007 & .012 & .029 & .011 & .018 & 26 & 67 & 4 \\
GLM-5.3-Flash, ZS (A8c) & .018 & .009 & .031 & .031 & .005 & .026 & .005 & .009 & .017 & .008 & .008 & .019 & 3 & 24 & 3 \\
MiniMax-M3.1-Flash, ZS (A8d) & .050 & .012 & .069 & .042 & .014 & .073 & .011 & .008 & .025 & .018 & .008 & .033 & 21 & 19 & 12 \\
Kimi-VL-A3B, ZS & .006 & .009 & .017 & .011 & .008 & .014 & .005 & .005 & .010 & .004 & .005 & .011 & 6 & 65 & 15 \\
Kimi-VL-A3B, MA & .007 & .004 & .007 & .012 & .005 & .013 & .003 & .003 & .002 & .003 & .004 & .005 & 25 & 90 & 33 \\
Kimi-VL-A3B, Seq & .010 & .014 & .031 & .007 & .012 & .033 & .009 & .014 & .016 & .005 & .012 & .021 & 1 & 27 & 5 \\
Kimi-VL-A3B, SW & .008 & .008 & .007 & .005 & .009 & .012 & .001 & .001 & .006 & .002 & .001 & .008 & 44 & 45 & 39 \\
Kimi-VL-A3B, CoT & .015 & .012 & .037 & .019 & .006 & .025 & .007 & .000 & .005 & .007 & .003 & .008 & 33 & 30 & 51 \\
Kimi-VL-A3B, MIPROv2 & .035 & .041 & .008 & .029 & .037 & .013 & .002 & .006 & .001 & .006 & .009 & .001 & 141 & 393 & 93 \\
Kimi-VL-A3B, MIPROv2-F1 & .033 & .036 & .063 & .034 & .032 & .055 & .011 & .007 & .015 & .008 & .007 & .014 & 110 & 434 & 79 \\
Kimi-VL-A3B, JudgeFit & .032 & .024 & .034 & .041 & .032 & .034 & .005 & .001 & .005 & .008 & .001 & .007 & 116 & 340 & 67 \\
Kimi-VL-A3B, VideoArgus & .010 & .017 & .050 & .007 & .019 & .046 & .005 & .009 & .038 & .003 & .009 & .030 & 30 & 53 & 20 \\
Kimi-VL-A3B, Harness$^2$ & .049 & .060 & .075 & .047 & .046 & .059 & .014 & .044 & .066 & .018 & .034 & .048 & 117 & 241 & 61 \\
Kimi-VL-A3B, LoRA & .017 & .032 & .029 & .013 & .029 & .026 & .006 & .008 & .006 & .005 & .011 & .006 & 52 & 152 & 49 \\
Kimi-VL-A3B, LoRA-F1 & .043 & .050 & .045 & .030 & .042 & .047 & .022 & .025 & .024 & .022 & .030 & .031 & 23 & 46 & 20 \\
Kimi-VL-A3B, JudgeFit-F1 & .092 & .063 & .075 & .104 & .067 & .084 & .010 & .003 & .011 & .015 & .004 & .017 & 344 & 929 & 190 \\
InternVL3.5-8B, ZS & .013 & .004 & .007 & .014 & .001 & .011 & .005 & .004 & .008 & .006 & .002 & .007 & 2 & 18 & 4 \\
InternVL3.5-8B, MA & .022 & .005 & .002 & .027 & .005 & .006 & .003 & .002 & .005 & .005 & .002 & .006 & 17 & 40 & 15 \\
Gemma 4 E4B, ZS & .008 & .016 & .043 & .003 & .013 & .042 & .002 & .010 & .037 & .003 & .009 & .033 & 11 & 12 & 5 \\
Gemma 4 E4B, MA & .013 & .008 & .010 & .017 & .011 & .022 & .012 & .002 & .006 & .013 & .005 & .009 & 11 & 29 & 2 \\
\midrule
\multicolumn{16}{l}{\emph{\method (\orch orchestrator) across video models}} \\
\method, \vlm & .022 & .050 & .084 & .024 & .051 & .079 & .028 & .017 & .006 & .012 & .008 & .028 & 81 & 212 & 43 \\
\method, Kimi-VL-A3B & .045 & .029 & .018 & .048 & .033 & .023 & .052 & .050 & .084 & .028 & .008 & .028 & 65 & 150 & 31 \\
\method, InternVL3.5-8B & .094 & .096 & .124 & .090 & .080 & .106 & .014 & .012 & .034 & .039 & .039 & .055 & 101 & 223 & 49 \\
\method, Gemma 4 E4B & .167 & .123 & .124 & .144 & .108 & .111 & .052 & .037 & .056 & .080 & .064 & .071 & 90 & 233 & 41 \\
\method-Full, \vlm & .048 & .069 & .020 & .052 & .060 & .025 & .012 & .017 & .032 & .016 & .015 & .011 & 58 & 207 & 34 \\
\midrule
\multicolumn{16}{l}{\emph{\method with other orchestrators}} \\
\method (Qwen-orch), \vlm & .134 & .140 & .164 & .144 & .145 & .161 & .051 & .041 & .049 & .049 & .042 & .063 & 195 & 586 & 116 \\
\method-Full (Qwen-orch), \vlm & .081 & .128 & .113 & .116 & .153 & .104 & .065 & .037 & .021 & .008 & .026 & .080 & 127 & --- & 36 \\
\method (GLM 5.3 Flash), \vlm & .107 & .133 & .157 & .097 & .107 & .140 & .032 & .051 & .090 & .070 & .082 & .111 & 58 & 156 & 32 \\
\method (GLM 5.3 Flash), Kimi-VL-A3B & .069 & .015 & .047 & .059 & .012 & .036 & .146 & .111 & .111 & .082 & .026 & .034 & 41 & 148 & 31 \\
\method (GLM 5.3 Flash), InternVL3.5-8B & .033 & .030 & .036 & .024 & .022 & .034 & .035 & .028 & .033 & .030 & .019 & .017 & 48 & 139 & 28 \\
\method (GLM 5.3 Flash), Gemma 4 E4B & .033 & .048 & .044 & .026 & .038 & .037 & .019 & .074 & .059 & .027 & .048 & .051 & 14 & 7 & 20 \\
\method (MiniMax M3.1 Flash), \vlm & .109 & .089 & .130 & .095 & .078 & .100 & .050 & .044 & .062 & .060 & .049 & .071 & 115 & 251 & 48 \\
\method (MiniMax M3.1 Flash), Kimi-VL-A3B & .058 & .062 & .071 & .050 & .054 & .072 & .050 & .023 & .026 & .036 & .007 & .013 & 133 & 305 & 60 \\
\method (MiniMax M3.1 Flash), InternVL3.5-8B & .083 & .047 & .083 & .061 & .039 & .080 & .033 & .027 & .027 & .037 & .021 & .042 & 83 & 203 & 47 \\
\method (MiniMax M3.1 Flash), Gemma 4 E4B & .058 & .093 & .092 & .061 & .072 & .079 & .019 & .036 & .035 & .022 & .045 & .049 & 93 & 240 & 44 \\
\bottomrule
\end{tabular}}
\subcaption{Baselines and \method across video models.}
\end{subtable}

\medskip
\begin{subtable}{\textwidth}
\centering
\setlength{\tabcolsep}{6pt}
\begin{tabular}{lc}
\toprule
\method search design (\orch, \vlm) & Test F1 \\
\midrule
Single split, no merge & .031 \\
Multi split, no merge & .077 \\
Multi split and merge (\method) & .012 \\
\bottomrule
\end{tabular}
\subcaption{Search-design ablations of \cref{tab:ablation}: test F1.}
\end{subtable}
\caption{\textbf{Standard deviations across three runs.} We report sample standard deviations for (a) all metrics and splits in \cref{tab:suppl:results_full} and (b) test F1 in the search design ablation (\cref{tab:ablation}). Related results appear in \cref{tab:results,tab:suppl:vlms,tab:suppl:grid,tab:results_vlm}.}
\label{tab:suppl:sd}
\end{table*}

\begin{table*}[t]
\centering
\small
\setlength{\tabcolsep}{6pt}
\begin{tabular}{lcccc>{\columncolor{oodshade}}c>{\columncolor{oodshade}}c>{\columncolor{oodshade}}c>{\columncolor{oodshade}}c}
\toprule
 & \multicolumn{4}{c}{Test} & \multicolumn{4}{c}{OOD} \\
\cmidrule(lr){2-5}\cmidrule(lr){6-9}
Error category & Annotated & Matched & Missed & Recall & Annotated & Matched & Missed & Recall \\
\midrule
Prompt adherence    & 54 & 17 & 37 & 31.5\,\% & 191 & 53 & 138 & 27.7\,\% \\
Physical violation  & 45 & 19 & 26 & 42.2\,\% & 80 & 21 & 59 & 26.2\,\% \\
Appear / disappear  & 36 & 8 & 28 & 22.2\,\% & 92 & 12 & 80 & 13.0\,\% \\
Motion              & 24 & 7 & 17 & 29.2\,\% & 76 & 14 & 62 & 18.4\,\% \\
Anatomy / body pose & 28 & 4 & 24 & 14.3\,\% & 84 & 16 & 68 & 19.0\,\% \\
Logical             & 18 & 1 & 17 & 5.6\,\% & 57 & 10 & 47 & 17.5\,\% \\
\midrule
All                 & 205 & 56 & 149 & 27.3\,\% & 580 & 126 & 454 & 21.7\,\% \\
\bottomrule
\end{tabular}
\caption{\textbf{Matched and missed errors by \spotlight category.} Results across all six \spotlight's error categories for the harness in \cref{fig:overview} (\orch orchestrator, \vlm VLM) on 80 test and 200 LTX-2 OOD clips. Recall is the fraction of annotated errors matched under \spotlight's rule. Of 240 test predictions, 56 matched and 184 were unmatched. Of 600 OOD predictions, 126 matched and 474 were unmatched. Only annotations carry category labels, so unmatched predictions have no category.}
\label{tab:suppl:categories}
\end{table*}

\begin{table*}[t]
\centering
\small
\setlength{\tabcolsep}{6pt}
\begin{tabular}{lccccccc>{\columncolor{oodshade}}c}
\toprule
 & \multicolumn{7}{c}{F1 by error category} & \\
\cmidrule(lr){2-8}
Method (\vlm) & Adherence & Physical & Appear / disappear & Motion & Anatomy & Logical & All & pred/clip \\
 & (54) & (45) & (36) & (24) & (28) & (18) & (205) & \\
\midrule
ZS (A1) & .062 & .082 & .017 & .040 & .034 & .021 & .145 & 3.0 \\
MA (A2) & .040 & .040 & .019 & .018 & .011 & .008 & .114 & 13.8 \\
Seq (A2) & .065 & .071 & .030 & .020 & .020 & .016 & .139 & 3.9 \\
SW (A2) & .058 & .038 & .021 & .025 & .013 & .010 & .128 & 9.2 \\
CoT (A2) & .098 & .080 & .033 & .034 & .026 & .024 & .173 & 3.3 \\
\addlinespace[2pt]
LoRA (A3) & .091 & .050 & .027 & .032 & .013 & .014 & .135 & 3.3 \\
LoRA-F1 (A3) & .107 & .058 & .029 & .039 & .014 & .017 & .153 & 3.1 \\
MIPROv2 (A4) & .056 & .058 & .030 & .016 & .019 & .011 & .122 & 4.2 \\
MIPROv2-F1 (A4) & .083 & .056 & .030 & .029 & .025 & .023 & .147 & 3.6 \\
\addlinespace[2pt]
JudgeFit (A11) & .035 & .036 & .020 & .016 & .016 & .007 & .111 & 18.3 \\
JudgeFit-F1 (A11) & .041 & .044 & .030 & .022 & .022 & .008 & .133 & 11.8 \\
VideoArgus (A12) & .096 & .035 & .012 & .029 & .010 & .007 & .107 & 2.5 \\
Harness$^2$ (A13) & .039 & .077 & .041 & .051 & .020 & .007 & .104 & 1.9 \\
\cmidrule{1-9}
\method (Deepseek-orch) & .113 & .118 & .062 & .054 & .045 & .023 & .239 & 3.9 \\
\bottomrule
\end{tabular}
\caption{\textbf{F1 by \spotlight error category.} Results for \method and the baselines in \cref{tab:results} on the 80-clip test split, across six categories (annotation counts in brackets). For each category, matched and unmatched annotations count as true positives and false negatives, respectively; all unmatched predictions count as false positives. Values are averaged over three runs; pred/clip denotes mean predictions per clip. \textbf{\method achieves the highest F1 in five of six categories and overall.} Counts for the harness in \cref{fig:overview} appear in \cref{tab:suppl:categories}.}
\label{tab:suppl:categories_baselines}
\end{table*}

\section{Ablation experiment details}
\label{sec:suppl:tables}
\label{sec:suppl:budgets}
The three search designs in \cref{tab:ablation} share the \spotlight-50 training pool, \orch orchestrator, \vlm VLM, \judge as judge, and seed. Each is run three times, reporting test F1 for the harness selected on held-out clips hidden from the orchestrator. The designs differ in training splits, candidate generation, and search budget (120--160 harnesses).

\paragraph{Single split, no merge.}
A single search uses 50 clips for training, with annotations visible to the orchestrator, 24-clip validation set $V$ outside the training set for ranking. Each of eight rounds comprises one call proposing 20 distinct plans and 20 separate implementation calls ($8 \times 20 = 160$ harnesses). From round two, these calls receive the memory of all previous candidates and the best harness so far in full. Each round proposes new candidates without further training subsets, iterative refinement, or merging. Every candidate is scored on both sets. After the final round, the harness with the highest held-out F1 is evaluated once on the 80-clip test split, excluded from search, yielding the result in \cref{tab:ablation}.

\paragraph{Multi split, no merge.}
This variant disables merging in \method and retains all other \spotlight-50 settings (\cref{tab:suppl:hparams}). Each of two rounds randomly partitions the 50 training clips into five disjoint subsets of 10. Per subset, the orchestrator generates three initial harnesses and refines each independently through three critique-and-rewrite steps, producing $2 \times 5 \times 3 \times (1+3) = 120$ harnesses. Each final candidate is scored on the 24-clip validation set $V$ outside the training set. The seed, validation set, and partitions match the \method runs. The candidate with the highest validation score is selected and evaluated once on the test split.

\paragraph{Multi split and merge (\method).}
The full method adds two merge iterations per round, each generating six harnesses that combine finalists from different subsets. The two best retained merges by validation score enter the search memory for subsequent rounds. This yields $2 \times (5 \times 3 \times 4 + 2 \times 6) = 144$ harnesses, adding 24 merge candidates to the previous design. All other settings are unchanged. The harness with the highest score on $V$ is selected and evaluated once on the test split.

\section{Orchestrator and VLM grid experiment setup}
\label{sec:suppl:extended}
\Cref{tab:suppl:vlms,tab:suppl:grid} evaluate \method across orchestrators and VLMs. Each grid cell averages three runs with the \spotlight-50 budget (\cref{sec:method:instantiation}), the same seed, and \judge as the judge, varying only the orchestrator and the VLM within the harness. \Cref{tab:suppl:vlms} compares \method with the benchmark's zero-shot and MA prompts for each VLM. Kimi-VL runs use two GPUs, one for the VLM and one for the tools. The Qwen-orch rows of \cref{tab:results} are a separate configuration and are not part of this grid (\cref{sec:suppl:qwen_orch}).

\section{IPV-Bench: transfer and in-domain optimisation experiments}
\label{sec:suppl:ipv}
IPV-Bench~\cite{bai2025impossible} contains 902 generated clips of impossible events, each annotated with a free-text answer and three paraphrases but no timestamps, and 650 real clips. We evaluate both OOD transfer of methods developed on \spotlight and in-domain optimisation of \method and trainable baselines. \Cref{tab:ipv} reports both settings, setup details below.

\paragraph{Setup.}
Three of the 902 generated clips have no reference answer in the benchmark’s annotation file and are excluded; the remaining 899 clips are partitioned by generator family into 400 training clips (Hailuo, Kling, Mochi, and Tongyi) and 499 test clips from the remaining families, including Hunyuan, Sora, Luma, and Pyramid Flow. We sample 400 of the 650 real clips once for false-alarm evaluation. All methods receive an empty prompt and rely on the video alone. We use \spotlight's evaluator with \judge, scoring only the explanations. A prediction matches the reference answer if its semantic score $S$, averaged over the four answer phrasings, is at least $0.7$. $\Mpct$ is the fraction of impossible clips with a match. Temporal segments are required but not scored. Every prediction on a real clip is a false alarm, so we report the fraction of real clips receiving at least one prediction.

\paragraph{F1$^*$ on IPV-Bench.}
\label{sec:suppl:ipv_f1}
Each impossible clip has one annotated event, described by an answer and three paraphrases, while real clips have none. The judge scores each prediction independently, assigning a semantic score $S$ averaged over the four phrasings. Recall is $\Mpct$, the fraction of impossible clips with at least one prediction satisfying $S \ge 0.7$. Precision pools all predictions across the evaluation set. Let $N_{\text{pred}}$ be the number of predictions on all clips, impossible and real, and $N_{S \ge 0.7}$ the number of predictions on impossible clips with $S \ge 0.7$. Then
\begin{align*}
\mathrm{Prec}^* &= \frac{N_{S \ge 0.7}}{N_{\text{pred}}}, \\[4pt]
\mathrm{F1}^* &= \frac{2\,\mathrm{Prec}^*\,\Mpct}{\mathrm{Prec}^* + \Mpct}.
\end{align*}
Every prediction on a real clip is therefore incorrect. If two of seven predictions on an impossible clip meet the threshold, both count towards precision, while the clip contributes once to recall. One-to-one detection precision would count only the Hungarian match.

Unlike the per-clip detection F1 in \cref{tab:results}, F1$^*$ pools prediction counts, credits multiple explanations that match the same event, and penalises every false alarm on real clips. Per-clip F1 instead assigns zero to a real clip with any predictions and excludes a silent real clip. For methods producing at most one prediction per clip, F1$^*$ equals pooled detection F1. Both \cref{tab:ipv,tab:suppl:ipv_full} evaluate all 899 test clips, with the latter additionally reporting $\Mpct$, $\mathrm{Prec}^*$, and per-clip detection F1 for reference.

\begin{table*}[t]
\centering
\small
\setlength{\tabcolsep}{3pt}
\resizebox{\linewidth}{!}{%
\begin{tabular}{lccccccc}
\toprule
 & \multicolumn{5}{c}{all test clips (899)} & \multicolumn{2}{c}{real clips (400)} \\
\cmidrule(lr){2-6}\cmidrule(lr){7-8}
Method (\vlm) & $\Mpct\,\uparrow$ & Prec$^*$\,$\uparrow$ & F1$^*$\,$\uparrow$ & F1\,$\uparrow$ & pred./clip & false alarms\,$\downarrow$ & pred./clip\,$\downarrow$ \\
\midrule
\multicolumn{8}{l}{\emph{Fixed prompts}} \\
ZS (A1)                     & 51.5 & .149 & .231 & .166 & 2.9 & 99.0\,\% & 2.8 \\
MA (A2)                     & \textbf{72.3} & .182 & .291 & .132 & 7.7 & 72.0\,\% & 5.7 \\
Seq (A2)                    & 45.9 & .131 & .204 & .143 & 2.8 & 99.5\,\% & 2.8 \\
SW (A2)                     & 65.1 & .107 & .183 & .116 & 6.5 & 100\,\% & 6.7 \\
CoT (A2)                    & 37.1 & .104 & .162 & .118 & 2.9 & 86.5\,\% & 2.5 \\
Qwen3-VL-30B ZS (A7)        & 63.9 & .119 & .200 & .109 & 8.1 & 100\,\% & 8.3 \\
\midrule
\multicolumn{8}{l}{\emph{Transferred from \spotlight, run as is}} \\
LoRA (A3)                   & 34.5 & .111 & .168 & .137 & 2.1 & 100\,\% & 2.1 \\
MIPROv2 (A4)                & 54.5 & .121 & .198 & .126 & 5.2 & 100\,\% & 5.1 \\
\method (Ours)              & 52.8 & .145 & .227 & .138 & 3.5 & 99.5\,\% & 3.4 \\
\midrule
\multicolumn{8}{l}{\emph{Optimised on the IPV pool}} \\
LoRA (A3)                   & 62.0\sd{1.0} & \textbf{.559}\sd{.022} & \textbf{.588}\sd{.014} & \textbf{.553}\sd{.017} & 0.6 & 15.2\sd{4.0}\,\% & 0.2 \\
MIPROv2 (A4)                & 51.3\sd{4.3} & .385\sd{.028} & .439\sd{.012} & .372\sd{.019} & 0.7 & 47.8\sd{21.1}\,\% & 0.5 \\
\method (Ours)              & 49.2\sd{5.7} & .551\sd{.048} & .520\sd{.053} & .463\sd{.043} & \textbf{0.5} & \textbf{7.5}\sd{3.3}\,\% & \textbf{0.1} \\
\bottomrule
\end{tabular}}
\caption{\textbf{Full IPV-Bench metrics.} We report $\mathrm{Prec}^*$, F1$^*$, and predictions per clip over all 899 test clips (\cref{tab:ipv,sec:suppl:ipv_f1}). Detection F1 uses one-to-one matching and is averaged per clip as in \cref{tab:results}. $\Mpct$ is the fraction of 499 impossible clips with a match. On the 400 real clips, we report false-alarm rate and predictions per clip. In-domain \method, MIPROv2, and LoRA each average three runs (LoRA seeds 8--10). Bold marks the unique best value in each column.}
\label{tab:suppl:ipv_full}
\end{table*}

\paragraph{Out-of-distribution transfer.}
We evaluate the \spotlight-trained LoRA adapter, MIPROv2-optimised prompt, and \method harnesses unchanged on the IPV-Bench impossible and real test clips. Transferred methods may produce multiple predictions per clip, with any matching prediction sufficient to count the clip as matched. \method matches 52.8\,\% of impossible clips, close to zero-shot (51.5\,\%), while averaging 3.5 predictions per clip. Like the other transferred methods, it flags nearly every real clip (99.5\,\%), consistent with the lack of an incentive to abstain when optimising only on clips containing errors.

\paragraph{In-benchmark optimisation.}
All real training and validation clips are sampled from the 250 clips outside the false-alarm evaluation set. LoRA follows its \spotlight recipe on the IPV pool plus 32 real clips, using an empty list as their target. The checkpoint with the highest detection F1 on a 48-clip validation subset (40 impossible, 8 real) is retained (\cref{sec:suppl:baselines}). MIPROv2 uses the pool plus 40 real clips, \method's task description, and at most one prediction per clip. Its per-clip objective is detection F1 on impossible clips and 1 for a real clip with no prediction, otherwise 0.

\method follows the \spotlight-50 configuration (\cref{tab:suppl:hparams}) with three adaptations. First, training includes 50 impossible clips sampled in proportion to generator family and 12 real clips. Validation comprises 24 impossible clips stratified by taxonomy label (physical, biological, social, and geographical laws) in the pool's proportions, plus 6 real clips disjoint from training. Second, the task description permits at most one error per clip, with an empty list indicating a plausible clip. Third, the objective averages per-clip detection F1, assigning 1 to an impossible clip with a matching prediction and 0 otherwise. Real clips receiving a prediction score 0, while those receiving none are excluded from the mean. False alarms thus reduce the search objective, as they also do under LoRA's training targets and MIPROv2's objective. Each selected harness is frozen and evaluated once on the test and false-alarm sets.

Over three runs, \method matches $49.2 \pm 5.7$\,\% of impossible test clips and flags $7.5 \pm 3.3$\,\% of real clips. Its mean false-alarm rate is lower than those of MIPROv2 ($47.8 \pm 21.1$\,\%) and LoRA ($15.2 \pm 4.0$\,\%), while its $\Mpct$ is approximately 13 percentage points below LoRA's 62.0\,\% (\cref{tab:suppl:ipv_full}). Including real clips during optimisation reduces false alarms across all three methods, with \method achieving the lowest false alarm rate.
\paragraph{Matched training data.}
In \cref{tab:ipv}, LoRA and MIPROv2 use all 400 impossible training clips plus real clips, while \method uses 50 impossible and 12 real clips. \Cref{tab:suppl:ipv_subset} repeats both baselines on \method's 62 training clips over three runs each (LoRA seeds 8,9,10), with all other settings unchanged, giving all three methods identical training data. 

\begin{table*}[t]
\begin{minipage}[t]{0.44\linewidth}
\centering
\small
\setlength{\tabcolsep}{4pt}
\begin{tabular}[t]{lccc}
\toprule
Method (\vlm) & F1$^*$\,$\uparrow$ & $\Mpct\,\uparrow$ & false alarms\,$\downarrow$ \\
\midrule
LoRA (A3)       & \textbf{.577}\sd{.023} & \textbf{59.2}\sd{2.0} & 15.9\sd{13.2}\,\% \\
MIPROv2 (A4)    & .438\sd{.030} & 43.5\sd{3.7} & 21.9\sd{12.7}\,\% \\
\method (Ours)  & .520\sd{.053} & 49.2\sd{5.7} & \textbf{7.5}\sd{3.3}\,\% \\
\bottomrule
\end{tabular}
\caption{\textbf{IPV-Bench with matched training data.} LoRA and MIPROv2 each use the same 50 impossible and 12 real training clips as \method. All methods are evaluated on the 899 test clips (\cref{tab:ipv}). The \method row is from \cref{tab:suppl:ipv_full}. With the full pool, LoRA and MIPROv2 reach $0.588 \pm 0.014$ and $0.439 \pm 0.012$ F1$^*$, respectively. Bold marks the best value per column.}
\label{tab:suppl:ipv_subset}
\end{minipage}\hfill
\begin{minipage}[t]{0.53\linewidth}
\centering
\small
\setlength{\tabcolsep}{4pt}
\resizebox{\linewidth}{!}{%
\begin{tabular}[t]{lccccc}
\toprule
 & no pred. & matched & matched, & words & ROUGE-L \\
Method (\vlm) & (\%) & (\%) & answered (\%) & per pred. & vs.\ annot. \\
\midrule
ZS (A1)                        & 0.0  & 51.5 & 51.5 & 15.7 & .246 \\
LoRA (A3), whole pool          & 1.1  & 62.0 & 62.7 & 19.0 & .334 \\
LoRA (A3), \method's 62 clips  & 7.2  & 59.2 & 63.8 & 18.1 & .311 \\
\method (Ours)                 & 17.0 & 49.2 & 59.3 & 28.8 & .238 \\
\midrule
Annotations                    & ---  & ---  & ---  & 19.3 & --- \\
\bottomrule
\end{tabular}}
\caption{\textbf{Predictions on 499 impossible IPV-Bench test clips.} \emph{No pred.} is the percentage of clips with no prediction. \emph{Matched} is the percentage with an annotation match under \judge, reported for all clips and for answered clips only. Words per prediction measures mean explanation length. ROUGE-L measures word-sequence overlap with the annotation. Results for both LoRA rows and \method are pooled over three runs each.}
\label{tab:suppl:ipv_outputs}
\end{minipage}
\end{table*}

\paragraph{LoRA and \method outputs.}
\Cref{tab:suppl:ipv_outputs} details the main-paper comparison. LoRA's explanations have the highest lexical overlap with the annotations, including when trained on 62 clips used to train \method. In this setting, it answers 93\,\% of impossible clips, with 64\,\% of those answers matching the annotations.

\section{Prediction flooding under \spotlight's reported score}
\label{sec:suppl:flood}

HSM$^*$ (\cref{tab:results_vlm}) replaces detection F1 with \spotlight's reported score $(M+\Mpct)/2$ as the search objective. All other settings match the \spotlight-50 configuration of \method, including the 50 training clips, validation and clean sets, splits, \orch orchestrator, and \vlm VLM. The reported score replaces F1 in refinement acceptance, merge retention, validation ranking, final selection, and the orchestrator's prompts.

Overprediction escalates within two rounds. The 60 refinement candidates in round 1 produce a median of 4.0 predictions per clip, with only one exceeding 40. The eight evaluated merge candidates already produce 86--187 predictions per clip, with training scores of 0.71--0.85. In round 2, 57 of the 60 refinement candidates exceed 40 predictions per clip, with a median of 145. Four of the five evaluated merge candidates then produce 1{,}221--1{,}341 predictions per clip, with training scores up to 0.884. We stop the search during this merge stage, after approximately 130 of 144 harnesses, and select the harness with the highest validation score (0.877).

On the test split, the selected harness produces 1{,}351 predictions per clip and achieves the highest reported score (0.824) and $\Mpct$ (0.840) in \cref{tab:results_vlm}, but precision and F1 fall to 0.002 and 0.003. It produces 1{,}376 predictions per clip on LTX-2. By comparison, \method optimised for F1 produces 3.9 predictions per test clip, with a reported score of 0.398 and F1 of 0.234.

Two early search designs that predate \method show the same behaviour across orchestrators. One uses Claude Opus~5.5 and the other \orch, each evaluated with both objectives. Under the reported score, the Opus search reaches a median of 52{,}225 predictions per clip by its third iteration. Under F1, no candidate's median exceeds five predictions per clip in either search.

\Cref{fig:tradeoff} illustrates the corresponding precision--coverage trade-off. Across its 19 methods, the reported score correlates strongly with predictions per clip on the test and LTX-2 splits (Spearman $\rho=0.87$ and $0.88$, respectively), whereas F1 shows much weaker correlations ($\rho=0.11$ and $0.25$).

\section{Qualitative analysis}
\label{sec:suppl:clip}
\begin{figure*}[t]
\centering
\includegraphics[width=\textwidth]{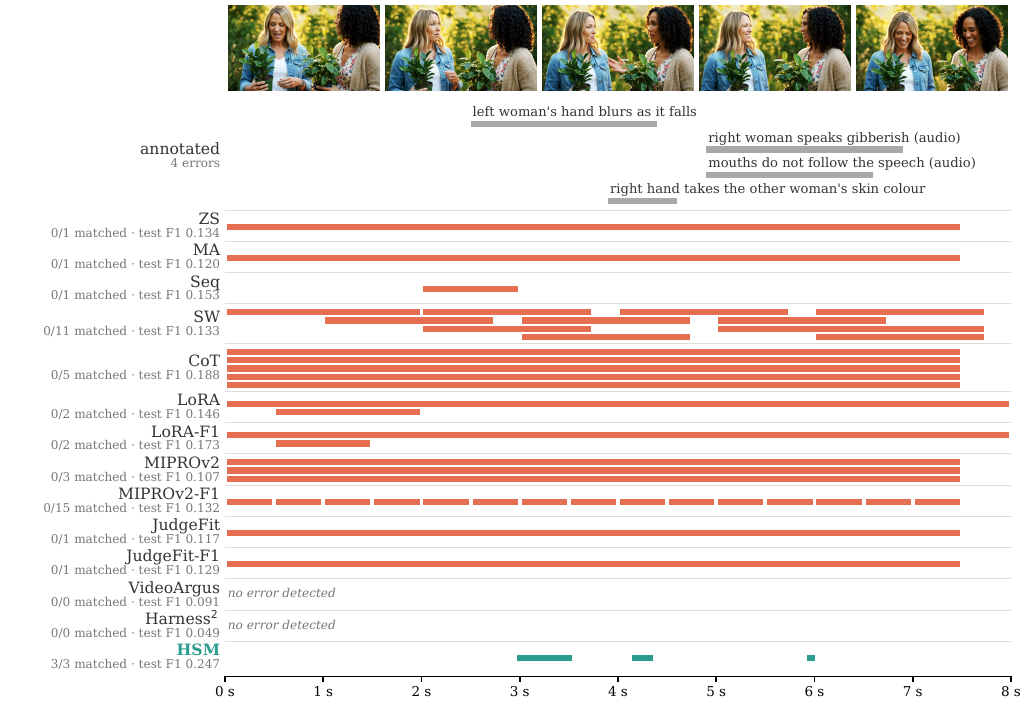}
\caption{\textbf{Best case: the Qwen3.5-9B baselines and \method on one test clip} (Veo~3, prompt 134). Annotated
errors in grey; predictions in teal (matched by the judge) or orange (not matched). Beside each name: predictions matched on this
clip, and the test F1 of the run shown over the 80 test clips.}
\label{fig:suppl:clip_qwen}
\end{figure*}

\begin{figure*}[t]
\centering
\includegraphics[width=\textwidth]{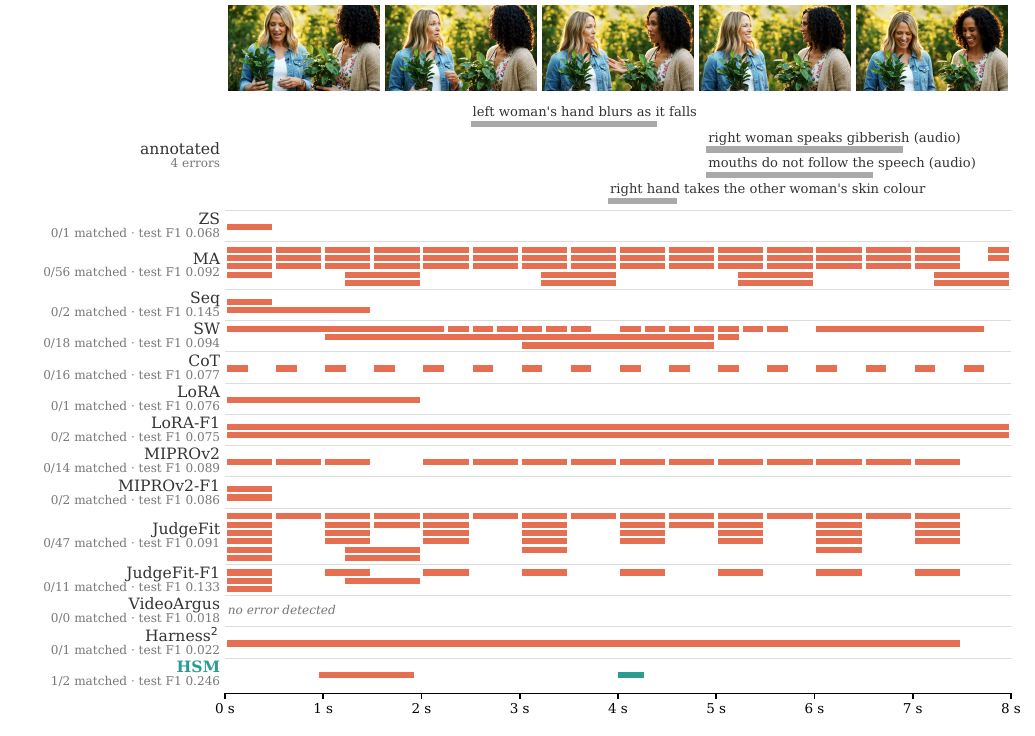}
\caption{\textbf{Best case: the Kimi-VL-A3B baselines and \method on the same clip.} Same format as
\cref{fig:suppl:clip_qwen}; rows with many overlapping predictions are drawn with thinner bars.}
\label{fig:suppl:clip_kimi}
\end{figure*}

\begin{figure*}[t]
\centering
\includegraphics[width=\textwidth]{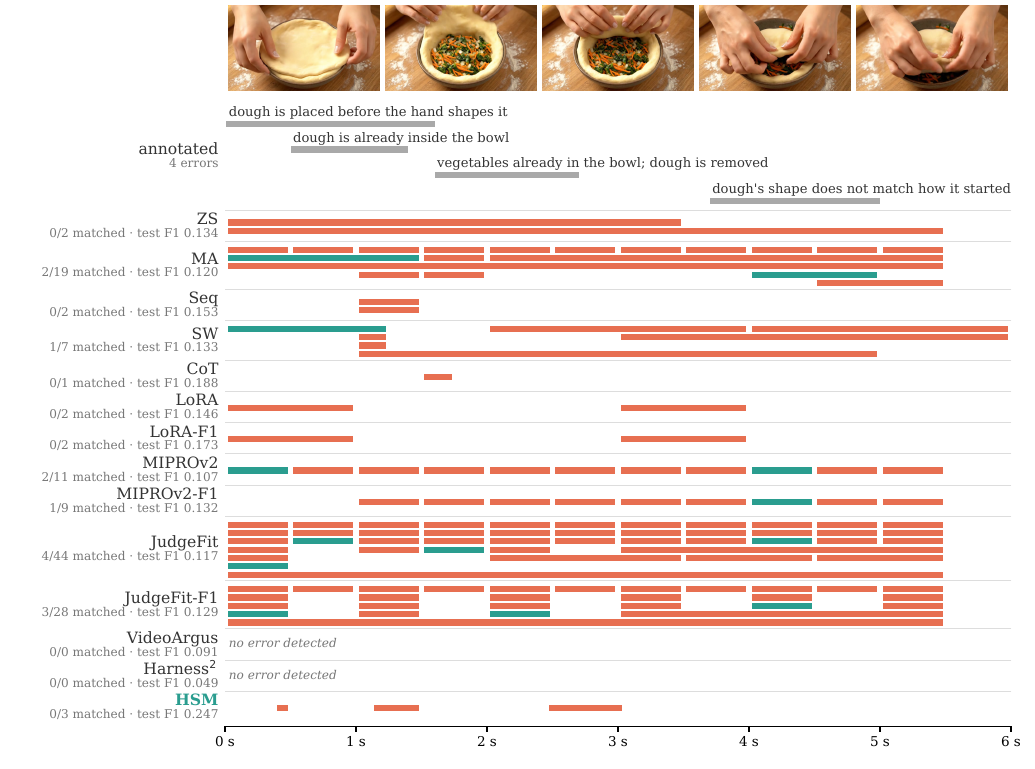}
\caption{\textbf{Worst case: the Qwen3.5-9B baselines and \method on the test clip where \method trails the
baselines most} (Veo~3, prompt 184). Same format as \cref{fig:suppl:clip_qwen}.}
\label{fig:suppl:clip_worst_qwen}
\end{figure*}

\begin{figure*}[t]
\centering
\includegraphics[width=\textwidth]{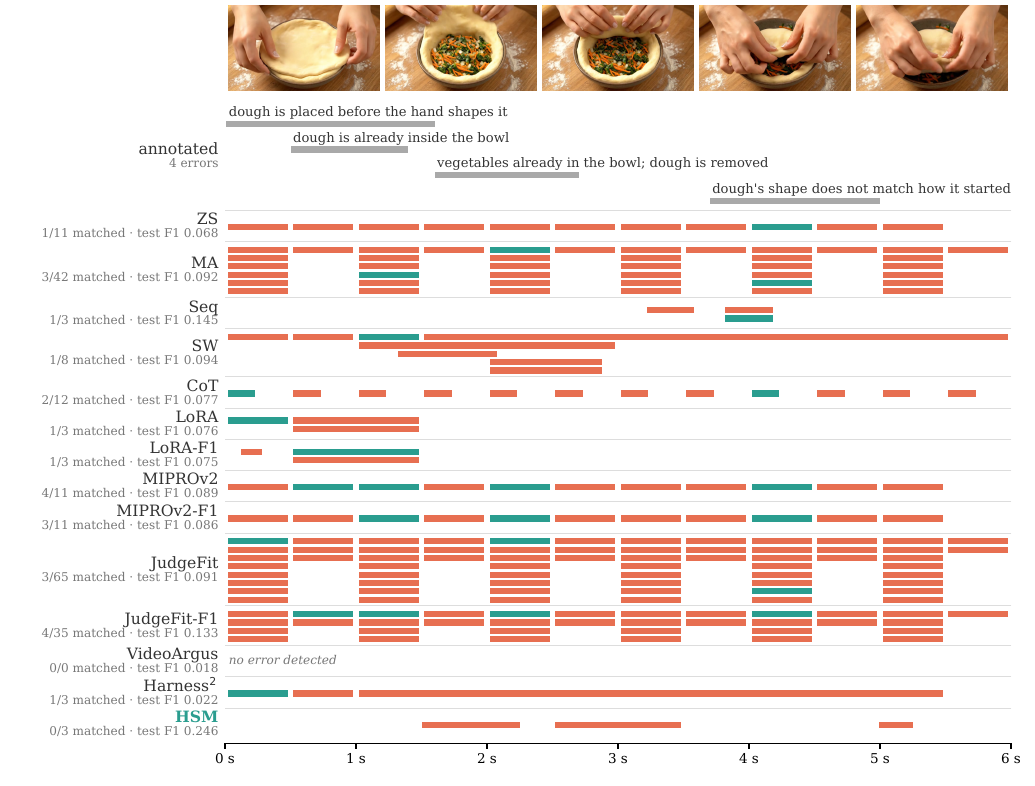}
\caption{\textbf{Worst case: the Kimi-VL-A3B baselines and \method on the same clip.} Same format as
\cref{fig:suppl:clip_qwen}.}
\label{fig:suppl:clip_worst_kimi}
\end{figure*}

\Cref{fig:suppl:clip_qwen,fig:suppl:clip_kimi,fig:suppl:clip_worst_qwen,fig:suppl:clip_worst_kimi} extend the timelines in \cref{fig:overview} to \method and the baselines in \cref{tab:suppl:results_full}, with one figure per VLM for two additional test clips. We select these clips by ranking the 55 test clips with two to four annotated errors, excluding the example in \cref{fig:overview}, by \method's per-clip F1 minus the baseline mean, summed across both VLMs. We show the highest- and lowest-ranked clips.

Each timeline shows one method's predictions, coloured teal if matched to an annotation and orange otherwise. Labels give the matched prediction count and the displayed run's F1 over all 80 test clips. Each row represents a single run, so its F1 may differ from the three-run means in \cref{tab:suppl:results_full}. For \method, we show the run with the highest test F1 among the three reported runs for each VLM: the Qwen3.5-9B harness in \cref{fig:overview} (0.247; mean 0.234) and a Kimi-VL-A3B harness scoring 0.246 (mean 0.219).

\emph{Best case} (Veo~3, prompt 134; \cref{fig:suppl:clip_qwen,fig:suppl:clip_kimi}). The clip contains four annotated errors. None of the baseline rows matches an error. \method matches three with three predictions using Qwen3.5-9B and one with two predictions using Kimi-VL-A3B.

\emph{Worst case} (Veo~3, prompt 184; \cref{fig:suppl:clip_worst_qwen,fig:suppl:clip_worst_kimi}). With \orch as orchestrator, \method produces three predictions with each VLM (Qwen3.5-9B and Kimi-VL-A3B), all unmatched. Qwen3.5-9B JudgeFit and Kimi-VL-A3B MIPROv2 and JudgeFit-F1 each match all four annotated errors, with 44, 11, and 35 total predictions, respectively.

\section{Configuration of \method}
\label{sec:suppl:hparams}
\Cref{tab:suppl:hparams} lists all parameters, caps, and budgets for the default \spotlight-50 configuration and the full-pool configuration (\cref{sec:results}). All \method results use \spotlight-50 unless stated otherwise. Notation follows \cref{sec:method,alg:hsm}.

\begin{table*}[t]
\centering
\small
\setlength{\tabcolsep}{4pt}
\begin{tabular}{lcc}
\toprule
 & \spotlight-50 budget & full pool \\
\midrule
\multicolumn{3}{l}{\emph{Search}} \\
training set $T$ & 50 videos & 200 videos \\
rounds $n$ & 2 & 3 \\
subsets per round $m$ (videos per subset) & 5 (10) & 10 (20) \\
lineages per subset $k$ & 3 & 3 \\
refinement steps per lineage $s$ & 3 & 3 \\
accept rule & \multicolumn{2}{c}{matched count not lower, F1 higher} \\
repair calls per failed candidate (create / rewrite / merge) & 1 / 2 / 1 & 1 / 2 / 1 \\
cross-subset evaluation: training videos outside a finalist's subset & 40 & 40 (sampled from 180) \\
finalists per round ($m k$) & 15 & 30 \\
finalists entering the search memory $a$ & all (up to 15) & all (up to 30) \\
merge iterations $u$ $\times$ proposals $q_m$ & $2 \times 6$ & $2 \times 6$ \\
finalists shown with code in the merge plan & 3 & 3 \\
kept merges per round $b$ (by $V$) & 2 & 2 \\
harnesses per run & 144 & 396 \\
\midrule
\multicolumn{3}{l}{\emph{Hidden sets (drawn outside the training set, never shown)}} \\
validation set $V$ & 24, drawn once & 80, drawn once \\
\quad stratification & \multicolumn{2}{c}{by error category; 10\,\% error-free; both generators} \\

\midrule
\multicolumn{3}{l}{\emph{Memory and orchestrator}} \\
records shown with code & 3 & 3 \\
context / output / thinking tokens & \multicolumn{2}{c}{262{,}144 / 16{,}384 / 8{,}192} \\
retries of an unusable answer & 1 & 1 \\
\midrule
\multicolumn{3}{l}{\emph{Candidate harness}} \\
frames per call (max) & 32 & 32 \\
VLM defaults: fps / frames / pixels / tokens / $T$ & \multicolumn{2}{c}{4 / 16 / 518{,}400 / 2{,}048 / 0} \\
tokens per VLM request (cap) & 32{,}768 & 32{,}768 \\
time per video / per candidate / smoke run & \multicolumn{2}{c}{600\,s / 6\,h / 900\,s} \\
tools & \multicolumn{2}{c}{RAFT, DINOv2, CLIP, Grounding DINO, SAM 2, CoTracker} \\
imports & \multicolumn{2}{c}{standard library, NumPy, PIL} \\
\midrule
\multicolumn{3}{l}{\emph{Models}} \\
orchestrator / VLM / judge & \multicolumn{2}{c}{\orch\ / \vlm\ / \judge} \\
\bottomrule
\end{tabular}
\caption{\textbf{\method parameters, caps and budgets.} The \spotlight-50 column gives the default settings and the full-pool column gives those for full-pool runs. IPV-Bench (\cref{sec:suppl:ipv}) uses the defaults with 62 training clips (50 impossible, 12 real), 30 validation clips (24 impossible, stratified by taxonomy label, and 6 real), and at most one prediction per clip. Validation scores rank and select the final harness. A merge is retained if it matches or exceeds its best parent's matched-error count and F1 on the round's training clips.}
\label{tab:suppl:hparams}
\end{table*}

\paragraph{Qwen-orch runs.}
\label{sec:suppl:qwen_orch}
In rows marked Qwen-orch, \vlm replaces both the \orch orchestrator and the \judge in-loop judge, making the search entirely local. The selected harness is then re-evaluated once with \judge for reporting. Settings follow \cref{tab:suppl:hparams}, with the orchestrator's output limit increased from 16{,}384 to 32{,}768 tokens to accommodate reasoning, which shares this limit rather than having a separate budget.

\section{\orch versus \vlm as orchestrator: recall by error category}
\label{sec:suppl:orch_categories}
\Cref{tab:suppl:categories_orch} compares the DeepSeek-orchestrator configuration, which uses gpt-4o-mini as the in-loop judge, with Qwen-orch, which uses Qwen3.5-9B for both orchestration and in-loop judging. Qwen-orch uses a much smaller model than \orch, with a larger output-token budget to accommodate its reasoning (\cref{sec:suppl:qwen_orch}). For each orchestrator, we select the highest-test-F1 run among three \spotlight-50 searches. The \orch harness appears in \cref{fig:overview,tab:suppl:categories}. Both harnesses use \vlm as the VLM and \judge for evaluation.

The Qwen-orch harness matches more annotated errors in all six categories on both splits. Overall recall increases from 27.3\,\% to 47.3\,\% on the test split and from 21.7\,\% to 47.2\,\% on LTX-2. On the test split, the largest gains are in logical errors (1 to 9 of 18) and anatomy errors (4 to 12 of 28), while motion changes least (7 to 8 of 24). On LTX-2 (OOD), the largest gain is in physical violations, from 26.2\,\% to 67.5\,\%.

\begin{table*}[t]
\centering
\small
\setlength{\tabcolsep}{3.5pt}
\begin{tabular}{lccccc>{\columncolor{oodshade}}c>{\columncolor{oodshade}}c>{\columncolor{oodshade}}c>{\columncolor{oodshade}}c>{\columncolor{oodshade}}c}
\toprule
 & \multicolumn{5}{c}{Test} & \multicolumn{5}{c}{OOD} \\
\cmidrule(lr){2-6}\cmidrule(lr){7-11}
 & & \multicolumn{2}{c}{\orch} & \multicolumn{2}{c}{Qwen-orch} & & \multicolumn{2}{>{\columncolor{oodshade}}c}{\orch} & \multicolumn{2}{>{\columncolor{oodshade}}c}{Qwen-orch} \\
\cmidrule(lr){3-4}\cmidrule(lr){5-6}\cmidrule(lr){8-9}\cmidrule(lr){10-11}
Error category & Annotated & Matched & Recall & Matched & Recall & Annotated & Matched & Recall & Matched & Recall \\
\midrule
Prompt adherence    & 54 & 17 & 31.5\,\% & 29 & 53.7\,\% & 191 & 53 & 27.7\,\% & 110 & 57.6\,\% \\
Physical violation  & 45 & 19 & 42.2\,\% & 23 & 51.1\,\% & 80 & 21 & 26.2\,\% & 54 & 67.5\,\% \\
Appear / disappear  & 36 & 8 & 22.2\,\% & 16 & 44.4\,\% & 92 & 12 & 13.0\,\% & 28 & 30.4\,\% \\
Motion              & 24 & 7 & 29.2\,\% & 8 & 33.3\,\% & 76 & 14 & 18.4\,\% & 34 & 44.7\,\% \\
Anatomy / body pose & 28 & 4 & 14.3\,\% & 12 & 42.9\,\% & 84 & 16 & 19.0\,\% & 29 & 34.5\,\% \\
Logical             & 18 & 1 & 5.6\,\% & 9 & 50.0\,\% & 57 & 10 & 17.5\,\% & 19 & 33.3\,\% \\
\midrule
All                 & 205 & 56 & 27.3\,\% & 97 & 47.3\,\% & 580 & 126 & 21.7\,\% & 274 & 47.2\,\% \\
\bottomrule
\end{tabular}
\caption{\textbf{Recall by \spotlight error category across two orchestrators.} For each orchestrator, \orch and \vlm (Qwen-orch), we select the harness with the highest test F1 among three \spotlight-50 searches and report recall on the 80 test and 200 LTX-2 OOD clips. Both harnesses use \vlm as the VLM. Recall is the fraction of annotated errors matched under \spotlight's matching rule.}
\label{tab:suppl:categories_orch}
\end{table*}

\section{Additional algorithm details}
\label{sec:suppl:algorithm}
\Cref{alg:hsm} specifies the procedure used for all \method results, with notation from \cref{sec:method} and parameters from \cref{tab:suppl:hparams}. We provide further implementation details below.

\paragraph{Information available to the orchestrator.}
Proposal calls receive the search memory $\mathcal{L}$ from completed rounds, containing compact records and the full code of the three highest-ranked candidates, together with archive-query results. They see the validation ranking but never the numerical scores or clips in $\mathcal{V}$. From round two, every proposal and merge call is preceded by an archive query, which can retrieve earlier code, per-video results, and refinement histories. Critique and rewrite calls access only their candidate's current harness, results on its training subset, and earlier attempts. They have no access to the global memory, other candidates' histories, or the archive. Merge proposals receive the round's finalists and their training F1 and cross-subset scores, the code of the top three finalists, a table showing which finalists match each annotated training error, and the merges generated so far.

\paragraph{Refinement acceptance.}
A rewrite replaces the current harness only if its matched-error count is no lower and its F1 is strictly higher on the subset's training clips. It may exchange one matched error for another if F1 improves, but cannot sacrifice matches for precision. The problem description in \cref{sec:suppl:prompts} still states an earlier, stricter rule, under which a rewrite must keep every error the current harness matched. The critique and rewrite prompts state the rule above, and acceptance is enforced in code, so the stricter sentence did not affect which rewrites were accepted.

\paragraph{Full-pool evaluation.}
With 200 training clips, each finalist is cross-evaluated on a fixed sample of 40 clips from the other subsets (\cref{tab:suppl:hparams}). A merge is therefore compared with its best parent on that parent's evaluation clips, comprising its training subset and cross-evaluation sample. In the \spotlight-50 configuration, this comparison covers all of $\mathcal{T}$. Training subsets and $\mathcal{V}$ keep both generators' clips for each prompt together. The test and LTX-2 splits (OOD set) are accessed only after search.

\section{Baseline setups}
\label{sec:suppl:baselines}
We run all baselines and score them as \method (\cref{sec:results:setup}). Unless a row specifies another model, \vlm is the VLM and \orch handles optimisation, clustering, and writing. The Kimi-VL-A3B block of \cref{tab:suppl:results_full} repeats the baselines with Kimi-VL-A3B-Instruct.

The \spotlight strategies (A1--A2, A6--A7) use the released prompts verbatim at 2\,fps with at most 32 frames; sliding-window inference uses 4\,fps over 2\,s windows. Gemini~2.5~Pro (A8), \orch (A8b), GLM~5.3~Flash (A8c), and MiniMax~M3.1~Flash (A8d) receive the zero-shot prompt and the same sampled frames as images.

\paragraph{LoRA (A3).}
LoRA~\cite{hu2021lora} fine-tunes the VLM on the annotations available to the search. From the 320-clip pool, we reserve 40 clips from 20 prompts, sampled with a fixed seed, for checkpoint selection. Inputs use the zero-shot prompt and frames, and targets are the requested JSON lists, with empty lists for error-free clips. Thus, A1 and A3 differ only in fine-tuning.

We apply LoRA to all LLM linear layers with rank 16 and alpha 32, freeze the vision tower, and train in bf16 with an effective batch size of 8 for 3--5 epochs. We sweep learning rates $5\times10^{-5}$, $10^{-4}$, and $2\times10^{-4}$, selecting the rate and epoch by the benchmark's scalar on the validation subset. The selected adapter is merged and served through vLLM. LoRA-F1 uses the same training runs but selects the rate and epoch by validation detection F1.

For Kimi-VL-A3B-Instruct, the same recipe and sweep use DeepSpeed ZeRO-3 on four GPUs; the selected checkpoint uses $2\times10^{-4}$ and three epochs. On IPV-Bench, training includes 32 real clips with empty-list targets, and checkpoint selection uses detection F1 on 40 impossible and 8 real clips (\cref{tab:ipv,sec:suppl:ipv}).

Each LoRA row reports three training runs with seeds 8, 9, and 10, controlling training and data shuffling. Each run repeats the full sweep and checkpoint selection on the same training, validation, and test clips, for both \spotlight VLMs and IPV-Bench.

\paragraph{MIPROv2 (A4).}
DSPy's MIPROv2~\cite{opsahl2024optimizing} optimises the zero-shot prompt using \vlm as the task model and \orch as the proposer, with no tools and the zero-shot frame settings. We use 57 instruction candidates and 85 trials on minibatches of 25 pool clips, following DSPy's trial-budget rule. Every five trials, DSPy re-evaluates the best candidate on the full validation set. We use no bootstrapped or labelled demonstrations because the inputs are frame sets. The objective is the benchmark's scalar on the pool. The selected prompt is frozen and evaluated once on the test and OOD splits.

MIPROv2-F1 repeats the \spotlight search with detection F1 as the objective, keeping the models, frames, and trial budget fixed. Each clip with annotated errors contributes its per-clip F1, giving the same macro-F1 objective as \method. For IPV-Bench, we optimise a separate prompt on the pool plus 40 real clips using 40 candidates, 60 trials, and the per-clip objective in \cref{sec:suppl:ipv}.

\paragraph{JudgeFit-loc (A11).}
JudgeFit~\cite{cao2026each} edits a judge's error taxonomy using feedback from its predictions. We use this taxonomy as the category list in \spotlight's multi-agent prompt, with one call per category. Following JudgeFit's Table~1, the fixed-schema, seed, and refined settings use the benchmark's six categories, an initial taxonomy, and its edited version, respectively.

The seed taxonomy comes from \vlm's free-form error notes on every pool clip, clustered by \orch into 3--8 named dimensions. Each candidate applies one edit (drop, redefine, merge, add, split, or no-op), with two candidates per round for at most three rounds. An edit is kept only if the objective improves.

We replace JudgeFit's ridge calibrator and diagnosis with the benchmark's annotation matching. A dimension is uninformative when its predictions rarely match, unstable when its leave-one-out effect changes sign across five folds, and redundant when it matches the same annotations as another dimension. Residuals are annotated errors missed by all dimensions. We omit the specificity term because the benchmark does not score prediction categories.

The objective is the benchmark's scalar on the 320 pool clips, scored by \judge. With \vlm, no edit improves the seed, so we evaluate the seed taxonomy on the test split. Each final taxonomy is evaluated once on the test split.

JudgeFit-loc-F1 uses the same seed taxonomy and changes only the objective. Mean per-clip F1 replaces the scalar in edit acceptance, leave-one-out effects, residual ranking, and the editor's instructions.

\paragraph{VideoArgus-loc (A12).}
VideoArgus~\cite{videoargus} evaluates videos against a rubric written from the generation prompt, using vision tools to gather evidence and a VLM judge to score each criterion. We use the authors' released prompts, schema, 22 dimensions, and tool vocabulary, with \orch writing one rubric per test prompt.

Evidence plans run on our tool worker, with Grounding~DINO and SAM~2 replacing SAM~3, DINOv2 replacing DINOv3, and the authors' flicker probe used unchanged. Unavailable tools (OCR, depth, and aesthetics) return the authors' tool-failure evidence, and the judge uses the remaining evidence. Web search is disabled. We follow the original frame settings: 4\,fps with at most 24 frames for question and detection steps, 8\,fps for dense probes over the full clip, and 12 representative frames for the judge, with the longest side resized to 768\,px.

We add one localisation call for each hard or soft criterion scored below the failure threshold of 5. This call uses \vlm with the zero-shot frames to return a segment and explanation. Exact duplicates within each clip are removed. The method uses no labels and is evaluated once on the test and OOD splits.

\paragraph{Harness$^2$ (A13).}
Harness$^2$~\cite{xu2026harness} edits an agent's system prompt, guardrails, memory, tool guidance, skills, scripts, subagents, and plugins for each task without a verifier. Parallel editor sessions propose procedural and domain-specific edits, run them, and compare the results with a reference rollout. Composition sessions combine edits, a blind selector chooses a candidate, and a reflector records lessons from the executions for later sessions.

We use the authors' prompts unchanged and fill their two extension points with a task adapter for error localisation and an executor block. The executor converts the harness into a candidate module using the same VLM and tools as \method (\cref{sec:method}). Harness$^2$ edits the same object without access to labels. Its base harness $H_0$ is identical to A1's zero-shot call.

Each editor session is approximated by one call containing the authors' workspace and the clip's frames. We compare this with agentic sessions on 10 training clips (\cref{tab:suppl:h2abl}), using \orch, four file commands, and a 40-command budget. Agentic sessions edit as often during proposal but rarely during composition, record one-third as many lessons, cost twice as much, and achieve no better test F1. We therefore use single-call sessions.

Adaptation uses the 50 \spotlight-50 clips without labels, three parallel branches ($k=3$), and the lesson library. We add a final composition step to freeze one harness, since the original method restarts each task from the base harness. Adaptation rollouts use VLM temperature $T=1$, following the authors' stochastic executor. Editor calls also use \orch at $T=1$: the original implementation leaves the temperature unspecified, and its branches differ only through sampling. The frozen harness $H^\ast$ and base control $H_0$ are evaluated at $T=0$, as are the other methods.

Each frozen harness is evaluated once on the test and LTX-2 splits, and the tables report means over three adaptation runs. With both VLMs, $H^\ast$ achieves lower F1 than $H_0$ (\cref{tab:suppl:results_full}).

\begin{table*}[t]
\begin{minipage}[t]{0.47\linewidth}
\hrule height 0pt
\centering
\small
\setlength{\tabcolsep}{4pt}
\resizebox{\linewidth}{!}{%
\begin{tabular}{lcc}
\toprule
 & single call & agentic \\
\midrule
training clips & 10 & 10 \\
proposal sessions that edited the harness (of 60) & 60 & 58 \\
composition sessions that edited the harness (of 30) & 30 & 7 \\
turns / file commands per session & 1 / 0 & 8.3 / 33.6 \\
library entries (of 10) & 10 & 3 \\
editor calls / cost (USD) & 111 / 1.46 & 987 / 3.19 \\
minutes per training clip & 7.7 & 9.6 \\
frozen $H^\ast$ & 1 skill, 1 plugin & $= H_0$ (KEEP\_BASE) \\
test F1 of $H^\ast$ & 0.130 & 0.134 \\
test F1 of $H_0$ (control) & 0.130 & 0.134 \\
\bottomrule
\end{tabular}}
\caption{\textbf{Harness$^2$ session mode}, single-call sessions (the workspace reproduced in one message) against
agentic sessions (the editor reads and writes the workspace with file commands, at most 40 per session), on the first
10 clips of the \spotlight-50 subset with \vlm; both arms frozen by the closing composition and scored once on the test
split. A session ``edited the harness'' when its proposal differed from the harness it started from; the agentic
compositions that did not (23 of 30) read the pool up to their command budget or returned no edit, and were retried
and then replaced by the base, so the agentic arm adopted the base on 9 of 10 clips and its closing composition kept
$H_0$ with a written justification. $H_0$ is scored once per arm (the 0.004 difference is the video model's
run-to-run variation at $T = 0$).}
\label{tab:suppl:h2abl}
\end{minipage}\hfill
\begin{minipage}[t]{0.47\linewidth}
\hrule height 0pt
\centering
\footnotesize
\setlength{\tabcolsep}{3pt}
\begin{tabular}{l cc}
\toprule
Method & F1 [95\,\% CI] & \method $-$ row \\
\midrule
ZS (A1) & .141 [.093, .195] & \textbf{$+$.093 [$+$.040, $+$.143]} \\
MA (A2) & .130 [.101, .159] & \textbf{$+$.104 [$+$.068, $+$.140]} \\
Seq (A2) & .155 [.109, .204] & \textbf{$+$.079 [$+$.033, $+$.125]} \\
SW (A2) & .132 [.106, .160] & \textbf{$+$.102 [$+$.070, $+$.132]} \\
CoT (A2) & .173 [.128, .219] & \textbf{$+$.061 [$+$.016, $+$.106]} \\
\addlinespace[2pt]
LoRA (A3) & .136 [.102, .172] & \textbf{$+$.098 [$+$.060, $+$.137]} \\
MIPROv2 (A4) & .120 [.086, .157] & \textbf{$+$.114 [$+$.074, $+$.154]} \\
MIPROv2-F1 (A4) & .141 [.104, .179] & \textbf{$+$.093 [$+$.055, $+$.132]} \\
\addlinespace[2pt]
JudgeFit (A11) & .121 [.100, .144] & \textbf{$+$.113 [$+$.086, $+$.139]} \\
JudgeFit-F1 (A11) & .134 [.109, .159] & \textbf{$+$.100 [$+$.071, $+$.130]} \\
VideoArgus (A12) & .087 [.052, .128] & \textbf{$+$.147 [$+$.103, $+$.188]} \\
Harness$^2$ (A13) & .087 [.056, .121] & \textbf{$+$.147 [$+$.112, $+$.183]} \\
\addlinespace[2pt]
Qwen3-VL-8B, ZS (A6) & .119 [.076, .166] & \textbf{$+$.115 [$+$.061, $+$.167]} \\
Qwen3-VL-8B, MA (A6) & .164 [.131, .199] & \textbf{$+$.070 [$+$.032, $+$.108]} \\
Qwen3-VL-30B, ZS (A7) & .107 [.075, .142] & \textbf{$+$.127 [$+$.084, $+$.169]} \\
Gemini-2.5-Pro, ZS (A8) & .167 [.132, .203] & \textbf{$+$.067 [$+$.030, $+$.105]} \\
DeepSeek-V4.1, ZS (A8b) & .175 [.139, .212] & \textbf{$+$.059 [$+$.023, $+$.094]} \\
\midrule
\method (Ours) & .234 [.199, .269] & --- \\
\bottomrule
\end{tabular}
\caption{\textbf{Test F1 and paired differences from \method.} We report detection F1 under \judge and the paired difference (\method minus each method), using three-run means as in \cref{tab:results}. We compute 95\,\% bootstrap intervals from 10{,}000 resamples of the 80 test clips, using the same resamples for all methods. Bold marks differences whose intervals exclude zero.}
\label{tab:suppl:bootstrap_ci}
\end{minipage}
\end{table*}

\section{Confidence intervals on the test split}
\label{sec:suppl:bootstrap}
\Cref{tab:suppl:bootstrap_ci} quantifies uncertainty from the 80 test clips using 95\,\% bootstrap confidence intervals for each method's F1 and its paired difference from \method, resampling clips jointly across methods.

\section{VLM control experiment: removing and swapping the VLM}
\label{sec:suppl:vlm_control}
To assess dependence on VLM outputs, we rerun the three selected \method harnesses from \cref{tab:results}, changing only their VLM calls while keeping code and tools fixed (\cref{tab:suppl:vlm_control}). Replacing every response with either an empty string or the constant sentence ``No error is visible in these frames.'' causes all three harnesses to return no predictions on either split, yielding zero F1. Swapping the VLM for Kimi-VL-A3B-Instruct without a new search preserves execution and yields $0.154 \pm 0.071$ test F1 and $0.176 \pm 0.045$ OOD F1. Across the three runs, one harness is unaffected by the swap (test F1 0.224 to 0.222), while the other two drop by 21--65\,\% across the two splits.

\begin{table*}[t]
\centering
\small
\setlength{\tabcolsep}{8pt}
\begin{tabular}{lcccc}
\toprule
VLM calls & Run 1 & Run 2 & Run 3 & mean $\pm$ sd \\
\midrule
\multicolumn{5}{l}{\emph{Test}} \\
\vlm (as searched) & .231 & .224 & .247 & .234\sd{.012} \\
Removed, empty answer & .000 & .000 & .000 & .000 \\
Removed, constant answer & .000 & .000 & .000 & .000 \\
Swapped to Kimi-VL & .081 & .222 & .159 & .154\sd{.071} \\
\midrule
\multicolumn{5}{l}{\emph{OOD}} \\
\vlm (as searched) & .217 & .219 & .204 & .213\sd{.008} \\
Removed, empty answer & .000 & .000 & .000 & .000 \\
Removed, constant answer & .000 & .000 & .000 & .000 \\
Swapped to Kimi-VL & .140 & .226 & .162 & .176\sd{.045} \\
\bottomrule
\end{tabular}
\caption{\textbf{VLM control.} Detection F1 (gpt-4o-mini judge) of the three \method runs found with \vlm when their
VLM calls are replaced; harness code and tools are unchanged. Runs 1--3 are the harnesses selected by the three
independent \method searches of \cref{tab:results}, one per search.}
\label{tab:suppl:vlm_control}
\end{table*}

\section{Search and inference cost}
\label{sec:suppl:cost}
\Cref{tab:suppl:cost} expands the main-paper cost comparison. The selected harnesses average four tool calls per clip and use fewer VLM calls, fewer prompt tokens, and less time than VideoArgus. Costs vary across the three harnesses, which require 10, 38, and 15 VLM calls, 61k, 128k, and 62k prompt tokens, and 64, 119, and 87 seconds per clip, respectively.

GPU-hours cover one search or training run, including VLM serving and all tool calls. With \vlm, LoRA uses two GPUs and all other methods use one. API costs include the optimiser or orchestrator and the in-loop judge. For \method, GPU-hours are measured from the search completed without a restart; API costs are averaged over the three searches, and inference figures over their three selected harnesses. VideoArgus has no search stage; its API cost covers rubric calls on the test and OOD clips.

\clearpage
\onecolumn
\raggedbottom
\begin{table}[t]
\centering
\small
\setlength{\tabcolsep}{8pt}
\begin{tabular}{l rr rrrr}
\toprule
 & \multicolumn{2}{c}{search / training} & \multicolumn{4}{c}{inference per clip} \\
\cmidrule(lr){2-3}\cmidrule(lr){4-7}
Method (\vlm) & GPU-h & API (US\$) & VLM calls & tool calls & prompt tokens & seconds \\
\midrule
ZS (A1) & --- & --- & 1 & 0 & 6.9k & 9 \\
MA (A2) & --- & --- & 6 & 0 & 39.7k & 50 \\
Seq (A2) & --- & --- & 5 & 0 & 33.0k & 31 \\
SW (A2) & --- & --- & 42.4 & 0 & 30.2k & 44 \\
CoT (A2) & --- & --- & 1 & 0 & 7.1k & 16 \\
\addlinespace[2pt]
LoRA (A3) & 12.9 & --- & 1 & 0 & 6.9k & 9 \\
MIPROv2 (A4) & 5.3 & 2.45 & 1 & 0 & 7.0k & 15 \\
MIPROv2-F1 (A4) & 8.6 & 2.09 & 1 & 0 & 7.0k & 39 \\
\addlinespace[2pt]
JudgeFit (A11) & 2.7 & 1.89 & 8 & 0 & 53.1k & 68 \\
JudgeFit-F1 (A11) & 3.5 & 1.50 & 7 & 0 & 46.5k & 58 \\
VideoArgus (A12) & --- & 0.55 & 23.9 & n/r & 141.1k & 229 \\
Harness$^2$ (A13) & 4.1 & 8.39 & n/r & n/r & n/r & n/r \\
\midrule
\method (Ours) & 14.3 & 4.62 & 20.9 & 4.0 & 83.9k & 90 \\
\bottomrule
\end{tabular}
\caption{\textbf{Search and inference costs with \vlm.} Search and training costs are for one run; inference costs are per-clip averages over the 80 test clips. n/r: not recorded. Details in \cref{sec:suppl:cost}.}
\label{tab:suppl:cost}
\end{table}
\section{Prompts}
\label{sec:suppl:prompts}
This section gives the prompts of the \spotlight-50 \method runs with \orch. \Cref{tab:suppl:prompts} lists the orchestrator calls of \cref{alg:hsm}; the templates follow. Every call starts with the system prompt, which contains the problem description, followed by the template of its step. Fields in double braces, such as \texttt{\{\{ k \}\}}, are filled in at run time, and \texttt{\{\% if \%\}} blocks are included only when their condition holds. The prompts say \emph{split} for a subset $S_j$, \emph{head} for a lineage's current harness, \emph{global logbook} for the search memory $\mathcal{L}$ and \emph{X} for the cross-subset score. HSM$^*$ uses the same prompts with the reported score $(M+\Mpct)/2$ named in place of detection F1.

\begin{table}[!h]
\centering
\small
\setlength{\tabcolsep}{3pt}
\begin{tabular}{@{}p{0.12\linewidth}p{0.47\linewidth}p{0.16\linewidth}@{}}
\toprule
Call & Shown to the orchestrator & Returns \\
\midrule
Plan & search memory, the subset's training clips with annotations & $k$ plans \\
Create & its plan, all plans, search memory, archive results, the subset's clips & one harness \\
Reflect & the lineage's current harness, its results and earlier steps & a critique \\
Rewrite & the critique and the same lineage view & a revised harness \\
Merge plan & finalists, complementarity table, merges so far, archive results & $q_m$ merge plans \\
Merge & its merge plan, the parents, complementarity table & one merged harness \\
Archive query & the search memory as the next call will show it & 2--10 queries \\
Repair & the failed harness and its error & a corrected harness \\
\bottomrule
\end{tabular}
\caption{\textbf{Orchestrator calls of \method.} Archive queries precede the create calls from round 2 and every merge call. Validation scores are never shown in any call.}
\label{tab:suppl:prompts}
\end{table}

\noindent\begin{minipage}{\linewidth}
\noindent\textbf{System prompt.}
\begin{lstlisting}[style=prompt]
You are the orchestrator of an automated optimisation loop. Your job is to design and improve candidate programs that solve the problem described below, using evidence from evaluations instead of guesses. You never execute code yourself; the harness runs every candidate you write and reports the scores back to you. Be concrete, keep your reasoning tied to the numbers you are shown, and follow every output protocol exactly: the harness parses your answers mechanically and a malformed answer wastes an iteration.

{{ problem_doc }}
\end{lstlisting}
\end{minipage}\par\medskip

\noindent\textbf{Problem description.}
\begin{lstlisting}[style=prompt]
# The problem

You are optimising an automatic system that inspects short AI-generated videos (5 to 8 seconds) and reports their visual generation errors. For each video the system receives the video itself and the text prompt it was generated from, and must return a list of error reports. Each report is a JSON object with a time segment and a reason:

```json
[{"segment": [1.2, 2.8], "reason": "The dog's front legs blur into each other and a third leg appears."}]
```

Human annotators marked the errors in these videos with the same format: a segment in seconds and a free-text reason. A report is good when its segment covers the annotated segment tightly (the segment term rewards precision: overlap divided by the predicted length) and its reason describes the same situation as the annotator's reason, as judged by a language-model similarity judge. The exact scoring function is given below under "Global objective".

Annotators used six categories, which are not scored and need not be predicted: physical violations, objects that appear or disappear, implausible motion, failure to follow the prompt, logical inconsistencies, and anatomy or body pose errors. Videos come from several text-to-video generators; a few videos contain no error at all, for which the correct output is an empty list.

# What you build

A candidate is a Python module that defines one function:

```python
def predict(video, api) -> list[dict]:
    ...
```

`video` carries the prompt and basic properties (duration, frame rate, size); it never carries annotations. `api` gives you frames at any frame rate and time window, a vision-language model that answers questions about frames, and a set of vision tools (optical flow, feature-change curves, text-image similarity, open-vocabulary detection, segmentation masks, point tracking) whose outputs are JSON. You decide the strategy entirely: which frames to look at, what to ask, how to combine tool signals with the model's answers, how to turn evidence into segments and reasons. The API reference and the tool catalogue follow below.

A candidate must be deterministic given the same video (temperature is 0 everywhere), must finish within the time limit per video, and may import only the standard-library and numeric modules listed in the API reference. There is no budget on how many model or tool calls a candidate makes, but every call is measured and reported back to you.

# How the optimisation works

The search runs for exactly 2 rounds. In every round the training videos are cut into 5 disjoint random splits (a fresh partition each round), and the splits are worked one after the other. A split has 3 lineages. Each lineage is created once: you write 3 distinct plans in one answer, then each plan is implemented in its own call as a complete candidate, the lineage's first head. Then every lineage is refined for 3 step(s). A step is two calls: a critique of the head (what it misses on the split's training videos and why, and the changes that should raise its detection F1), then a rewrite of the head from that critique. The rewrite is tested on the split's training videos (annotations available to you for analysis, never to the candidate) and becomes the new head only if it still matches every annotated error the head matched and its F1 there is higher; otherwise the head stays and the rewrite is recorded as tried. The critique and the rewrite see their own lineage only: its head with its code and per-video results, and the steps tried so far. When the round's splits are done, every lineage's head are the round's finalists, and each is also scored on the other splits' training videos of the round (X: videos its lineage never saw). A merge step follows: in 2 merge iteration(s) you first write 6 merge plans at a time (no two plans of one iteration may combine the same parents), then implement each as one harness that combines complementary finalists (or earlier merges of the round). Merged harnesses are scored on all of the round's training videos (annotations available to you); a merge is kept only if it matches at least as many annotated errors as its best parent there and its F1 is not lower, and the top 2 kept merges join the finalists. Finalists and kept merges enter the global logbook, which is ordered by their score on a fixed validation set V of 24 videos outside the training data; V's numbers are never shown to you, nor those of the fresh clean set C of 12 videos per round that measures how far the order adapted to V. The global logbook shows each record's training F1 and, for finalists, X. The create calls of later rounds see the global logbook and the archive, which you can query from the second round on and before every merge call. After the last round the record with the highest V is the result; you are also asked once to pick the record you expect to do best on new videos. The result is evaluated once on held-out test sets that no step of this process ever reads.

# Dataset

- Benchmark: `spotlight`. Optimisation split: `early_subset` (early_subset), 50 videos from 25 prompts.
- Generators in the split: seedance (25), veo3 (25).
- Annotations: 126 annotated errors in total; 0 videos are annotated as error-free.
- Durations: 5.04 to 8.00 s.
- Prompt: every candidate receives the text prompt the video was generated from (`video.prompt`).
- The final test sets are held out and are never read by any step of this loop; no counts or contents of them are shown.

# Candidate API and vision tools

[Reference omitted here: the signatures, parameters and output fields of the six API methods (frames, vlm, tool, tool_names, parse_json, log) and of the six vision tools (raft, dinov2, clip, gdino, sam2, cotracker).]

# Hardware and time limits

- Vision-language model behind `api.vlm`: `Qwen/Qwen3.5-9B` (temperature 0.0, thinking off). Defaults when a candidate does not specify: 4 fps, 16 frames, 518400 pixels per frame, 2048 output tokens.
- At most 32 frames per VLM call and 32768 tokens per VLM request; frames are capped at 518400 pixels each. Up to 8 VLM requests run in flight inside one candidate evaluation.
- Vision tools enabled in this run: clip, cotracker, dinov2, gdino, raft, sam2; they run one at a time on a dedicated GPU and their outputs are cached per video and parameters.
- Time limits: 600 s per video, 21600 s per candidate on one video set, 900 s for the one-video smoke run every candidate passes before evaluation.
- Allowed imports in a candidate: json, math, re, statistics, itertools, collections, functools, typing, dataclasses, enum, random, copy, operator, string, heapq, bisect, abc, time, textwrap, difflib, logging, numpy, PIL.
- Whole run: wall-clock guard 36 h and API spend guard $30; when a guard trips, the current global iteration finishes and the loop stops.
- No budget on VLM or tool calls per candidate; every call is measured and reported in the logbook.

# Global objective

Formula id `spotlight_mean_M_Mpct`.

Per video, with the annotated errors G (n_gt of them) and the predicted reports R:
- for every pair (g, r): `P = overlap(g.segment, r.segment) / length(r.segment)` (0 without overlap; the segment term rewards precision, so a long segment that covers a short annotation scores low), `S` = the similarity judge's rating of the two reasons (an integer 0 to 10 divided by 10), and `score = (P + S) / 2`;
- a one-to-one (Hungarian) assignment of predictions to annotated errors that maximises the summed score;
- assigned pairs with `score >= 0.7` are kept; `M` = mean score of the kept pairs, `M% = kept / n_gt`; `M = M% = 0` when either list is empty or nothing is kept. Over a set of videos M and M% are aggregated as the macro mean over the videos that have at least one annotated error (videos annotated as error-free are left out of the mean): M and M% are first computed per video, then averaged. The global objective is detection F1, not the scalar `(M + M%) / 2`. A matched prediction is a kept pair as above. Per video: `Prec = matched / n_pred`, `Rec = matched / n_gt` (this is `M%`), `F1 = 2 Prec Rec / (Prec + Rec)` (0 when both are 0). A video with annotated errors and no prediction scores F1 = 0; a video annotated as error-free scores F1 = 0 if it has any prediction and is left out of the mean if it has none. The set's F1 is the mean of the per-video F1 over the videos where it is defined, error-free videos included. Every prediction that is not matched lowers precision, so extra or duplicate reports lower the objective; higher is better. The scalar, M and M% are still computed and shown, for reporting. The error `type` field is not scored.

The similarity judge is `gpt-4o-mini-2024-07-18` at temperature 0; candidates cannot call it.

# Loop hyperparameters and caps

- Rounds: exactly 2; no early stop. Splits per round: 5 disjoint random training splits, re-drawn every round.
- Per split: 3 lineages, created by one plan call (3 plans of about 200 words) and 3 implementation calls, then 3 refinement step(s) per lineage (a critique and a rewrite each); a rewrite replaces the head only if it keeps every annotated error the head matched and raises its F1 on the split's training videos.
- X: every finalist's F1 on the other splits' training videos of its round, shown in the global logbook.
- Merge step: 2 iteration(s) x 6 merged harnesses per round, scored on all of the round's training videos; no two plans of an iteration with the same parents; a merge is kept only if it is not below its best parent there; the top 2 kept merges enter the global logbook.
- Validation set V: 24 fixed videos outside the training data, never shown; it orders the global logbook and picks the result. Clean set C: 12 fresh videos per round, never shown, never reused.
- Candidates in all: 144.
- Failures: a created candidate gets 1 repair call(s), a rewrite 2; if it still fails it is recorded with score 0 (a failed rewrite leaves the head as it is).
- Your answers: at most 16384 output tokens per call (thinking included where the model thinks; an answer cut at this limit is unusable), about 8192 of them for thinking; every prompt fits within a 262144-token context together with the answer.
- Any answer outside these caps is rejected with the reason and you are asked once more.

# What you see

- You see the annotations (segments and reasons) of the current split's training videos in the split's calls, and of all the round's training videos in the merge step, in the video listings and next to every per-video score. The validation set's and the clean sets' annotations and scores are used for scoring only and never reach you, not even as numbers. Candidates never see any annotation: `predict` receives only the video, its prompt and the API.
- Every evaluation is reported per video (scalar, M, M%, Prec, Rec, F1, annotated and predicted counts) and per prediction (matched or not, P, S).
- You never see the vision-language model's raw answers; only the candidate's predictions and their scores.
- Plan and create calls show the global logbook as a table of every record (training F1, X, source, lineage) in V's order, the top 3 records with their code and the others as one line each; the critique and rewrite calls show their own lineage only (the head's code and per-video results, every step tried so far).
- A record "in full" in these views is its compact form: reasoning, a 60-word summary of the program (your <<<SUMMARY>>> section), a diff line against the previous record, scores, and per video the numbers only (GT errors, matched, missed, predictions, M, M%, Prec, F1, P_mean and S_mean of its predictions); the code is shown for the top 3 records by score only (the best record always).
- Every record of the run, including the per-objective records of earlier global iterations, is kept in an archive with its code and full per-video detail (predictions, the GT each matched or missed, P and S). From the second round on, before every created candidate, and before every merge plan and merged harness, before a call you are asked for 2 to 10 archive queries (tools: record, video_history, search, compare) on the logbook you are about to see; their results (up to about 65536 tokens) are added to that call.
\end{lstlisting}

\noindent\begin{minipage}{\linewidth}
\noindent\textbf{Plan.}
\begin{lstlisting}[style=prompt]
# Step: plans for round {{ global_iter }} of {{ n_rounds }} ({{ split_note.title }})

{{ split_note.text }}

Write {{ k }} distinct plans for new candidates. Each plan becomes one lineage of this split: it is implemented once as a complete candidate by a separate call that sees all {{ k }} plans, and that candidate is then refined on its own for
{{ steps }} step(s). A plan must be specific enough to implement on its own: which frames to look at (rate, window), what to ask the vision-language model and how to phrase it, which tool signals locate or confirm events, and how evidence becomes segments and reasons. About {{ plan_words }} words per plan.
{% if n_records > 0 %}The global logbook below holds the best candidates of the finished rounds. Plans may build on them (say which record and what changes) or try new approaches; make them genuinely different from each other, since each starts its own lineage.{% else %}Make the plans genuinely different from each other, since each starts its own lineage.{% endif %}

{% if n_records > 0 %}
{{ logbook_view }}
{% endif %}

## Training videos of this split ({{ n_train }})
{{ training_videos }}

Answer with one ```json block of this exact shape and nothing else outside it, with exactly {{ k }} plans:
```json
{"plans": [{"title": "<short name>", "approach": "<the plan>"}, ...]}
```
\end{lstlisting}
\end{minipage}\par\medskip

\noindent\begin{minipage}{\linewidth}
\noindent\textbf{Create.}
\begin{lstlisting}[style=prompt]
# Step: candidate {{ slot }} of {{ k }} (round {{ global_iter }} of {{ n_rounds }}, {{ split_note.title }}, lineage {{ slot }})

{{ split_note.text }}

Implement plan {{ slot }} below as a complete candidate: it is the start of lineage {{ slot }} of this split, which is then refined for {{ steps }} step(s). The other plans of this split are listed for context only: each is implemented by its own call and becomes its own lineage, so stay with your plan and do not copy another one. The candidate is scored on the training videos below; that F1 is its score.{% if n_records > 0 %} Read the global logbook as evidence and use it where your plan leaves details open{% if archive_results %}, with the archive results below{% endif %}. Do not repeat a candidate already in it.{% endif %}

## Your plan (plan {{ slot }}): {{ plan.title }}
{{ plan.approach }}

## All plans of this split
{% for p in plans %}{{ loop.index }}. **{{ p.title }}**: {{ p.approach }}
{% endfor %}
{% if n_records > 0 %}
{{ logbook_view }}

## Best global record (the first of the global logbook)
{{ best_record }}
{% endif %}{% if archive_results %}
## Archive results
What you asked the archive for before this call.

{{ archive_results }}
{% endif %}
## Training videos of this split ({{ n_train }})
{{ training_videos }}

{{ protocol }}
\end{lstlisting}
\end{minipage}\par\medskip

\noindent\begin{minipage}{\linewidth}
\noindent\textbf{Reflect.}
\begin{lstlisting}[style=prompt]
# Step: reflection on lineage {{ lineage }} of {{ k }} (round {{ global_iter }} of {{ n_rounds }}, {{ split_note.title }}, refinement step {{ step }} of {{ steps }})

{{ split_note.text }}

You are the critic of one lineage. Below is its current head (the candidate the lineage stands on), its predictions on this split's {{ n_train }} training videos next to the annotated errors, its scores, and the steps of the lineage so far, with the rewrites that were tried and rejected. The next call rewrites the head from your critique; it is accepted only if
{{ accept_text }}.

Write a critique aimed at detection F1 on these videos. F1 is the per-video harmonic mean of precision (the share of predictions that match an annotated error) and M% (the share of annotated errors matched), averaged over the videos; Prec, M%, the scalar and the per-video numbers are shown so you can say which of the two terms a change moves. Look for the patterns across videos rather than one video at a time: which kinds of error the head misses everywhere, where its unmatched predictions come from (segments placed wrongly: low P; descriptions that do not match: low S; predictions on stretches without an error), and what in the code causes it. A head that returns no predictions at all needs a different approach, not a parameter change. Do not propose again what the rejected rewrites already tried, unless you say why it failed and what is different now.

Use exactly these three headings, in this order, and nothing else outside them:

## What it misses
The error kinds and patterns it misses or gets wrong across the videos, each with the video ids that show it.

## Why
The mechanism in the code that causes each one.

## Changes
One or two concrete edits, ordered by the F1 you expect them to gain. For each: what to change in the code, which term of F1 it moves, and which missed errors it is for (when it adds predictions) or which unmatched predictions it removes (when it removes predictions).

{{ lineage_view }}

## Training videos of this split ({{ n_train }})
{{ training_videos }}
\end{lstlisting}
\end{minipage}\par\medskip

\noindent\begin{minipage}{\linewidth}
\noindent\textbf{Rewrite.}
\begin{lstlisting}[style=prompt]
# Step: rewrite of lineage {{ lineage }} of {{ k }} (round {{ global_iter }} of {{ n_rounds }}, {{ split_note.title }}, refinement step {{ step }} of {{ steps }})

{{ split_note.text }}

Rewrite the head of this lineage by applying the critique below. Change what the critique asks for and keep the rest of the head as it is: the rewrite replaces the head only if {{ accept_text }}; otherwise the head stays and your rewrite is recorded as tried. Return the complete module, not a patch.

## Critique
{{ reflection }}

{{ lineage_view }}

## Training videos of this split ({{ n_train }})
{{ training_videos }}

{{ protocol }}

Add a `<<<DIFF>>>` section right after `<<<REASONING>>>`: one line saying what this rewrite changes relative to the head (for example "VLM confirmation threshold 0.6 -> 0.4; dropped the duplicate-merge pass").
\end{lstlisting}
\end{minipage}\par\medskip

\noindent\begin{minipage}{\linewidth}
\noindent\textbf{Merge plan.}
\begin{lstlisting}[style=prompt]
# Step: merge plans for round {{ global_iter }} of {{ n_rounds }} (merge iteration {{ iteration }} of {{ n_iters }})

The {{ n_fins }} finalists below are the heads of this round's split lineages. Each lineage was refined on one split's training videos only, so they find different errors. Your task is to combine complementary finalists into merged harnesses that raise detection F1: more annotated errors matched (M%) without extra unmatched predictions (precision). A merge that simply concatenates its parents' predictions usually lowers precision: decide how overlapping or duplicate predictions become one, which parent is trusted for which kind of error, and how segments and reasons are chosen.

The complementarity table covers all {{ n_all }} training videos of the round: a finalist's numbers there join its own split and its X (the other splits, which its lineage never saw). Every merged harness is scored on these {{ n_all }} videos; that F1 is its score. A merge is kept only if, on its best parent's own videos (that parent's split and X), it matches at least as many annotated errors as the parent and its F1 there is not lower; of the kept merges, the top {{ kept }} by a validation score you do not see join the finalists in the global logbook. {% if n_merges > 0 %}The merges written so far in this round are listed with their results: refine the promising ones or try other combinations.{% endif %}

Write {{ q }} distinct merge plans. Each names its parents by label or short name (at least two: finalists{% if n_merges > 0 %} or merges of this round{% endif %}; you choose how many) and says, specifically enough to implement on its own, what each parent contributes, how their outputs are combined and deduplicated, and which evidence in the table supports the choice. **No two plans may have the same set of parents.** About {{ plan_words }} words per plan.

## Complementarity on the round's {{ n_all }} training videos
{{ complementarity }}

## Finalists ({{ n_fins }}, best X first; the top {{ code_top }} with code, the others' code is in the archive)
{{ finalists }}

## Merges of this round so far ({{ n_merges }})
{{ merges }}
{% if used_sets %}
Parent sets of earlier merge iterations (they may be used again): {{ used_sets }}
{% endif %}{% if archive_results %}
## Archive results
What you asked the archive for before this call.

{{ archive_results }}
{% endif %}
## Training videos of the round ({{ n_all }})
{{ all_videos }}

Answer with one ```json block of this exact shape and nothing else outside it, with exactly {{ q }} merge plans:
```json
{"merges": [{"title": "<short name>", "parents": ["<label or F-name>", "<label or F-name>"], "approach": "<the plan>"}, ...]}
```
\end{lstlisting}
\end{minipage}\par\medskip

\noindent\begin{minipage}{\linewidth}
\noindent\textbf{Merge.}
\begin{lstlisting}[style=prompt]
# Step: merge {{ slot }} of {{ q }} (round {{ global_iter }} of {{ n_rounds }}, merge iteration {{ iteration }} of {{ n_iters }})

Implement merge plan {{ slot }} below as one complete candidate that combines its parents. The other merge plans are listed for context only: each is implemented by its own call. The merged harness is scored on all {{ n_all }} training videos of the round; that F1 is its score. It is kept only if, on its best parent's own videos (that parent's split and X), it matches at least as many annotated errors as the parent and its F1 there is not lower. Keep what each parent does well, and make sure overlapping predictions of different parents become one prediction: detection F1 counts every unmatched prediction against precision. Do not repeat a candidate already shown.

## Your merge plan (merge {{ slot }}): {{ plan.title }}
Parents: {% for p in plan.parents %}`{{ p }}`{% if not loop.last %}, {% endif %}{% endfor %}

{{ plan.approach }}

## All merge plans of this iteration
{% for p in plans %}{{ loop.index }}. **{{ p.title }}** (parents {{ p.parents | join(", ") }}): {{ p.approach }}
{% endfor %}
## Parents
{{ parents }}

## Complementarity on the round's {{ n_all }} training videos
{{ complementarity }}

## Merges of this round so far ({{ n_merges }})
{{ merges }}
{% if archive_results %}
## Archive results
What you asked the archive for before this call.

{{ archive_results }}
{% endif %}
## Training videos of the round ({{ n_all }})
{{ all_videos }}

{{ protocol }}
\end{lstlisting}
\end{minipage}\par\medskip

\noindent\begin{minipage}{\linewidth}
\noindent\textbf{Archive query.}
\begin{lstlisting}[style=prompt]
# Step: archive queries before writing {{ what }}

Below is the compact logbook{% if best_record %} and the best record{% endif %} exactly as the next call will show them, where you write {{ what }}. The compact logbook shows every record's reasoning, summary, scores and per-video numbers, but the code of only the top {{ top_full }} records by score and no per-video detail (the predictions with their segments and reasons, the GT each one matched or missed, P and S per prediction). Every record of this run ({{ n_archive }} so far, including the per-objective records of earlier global iterations that the merged logbook drops) is in the archive.

After reading the logbook, the next step is to query the archive deeply for the interesting candidates from the history before writing the new candidate: the code of the records you may build on, the per-video detail of the records whose numbers you need to explain (which GT was missed, whether a miss is placement (low P) or description (low S)), the history of the videos that keep failing, and the comparison of close candidates. Ask for at least
{{ min_queries }} and at most {{ max_queries }} queries. Their results are added to the next call, up to about
{{ result_tokens }} tokens in total and {{ result_tokens_each }} per query, so ask for the parts you need rather than everything. You will not see the results before that call: ask for everything you want now.

## Archive tools
- `record`: one record. `{"tool": "record", "ref": "<label or candidate hash>", "parts": ["reasoning", "code", "detail", "error"]}`.
  `parts` is optional (default: all four). A label is `g<k>·<objective>·i<n>` as the logbook titles show it (`/` or `:` instead of `·` is accepted); a candidate hash returns every record holding that code.
- `video_history`: every candidate's predictions on one video, with the GT matched or missed and P / S per prediction.
  `{"tool": "video_history", "video_id": "<video id>", "limit": 20}`.
- `search`: full-text search over reasoning, summaries and code. `{"tool": "search", "text": "<words>", "k": 5}`.
- `compare`: two records: their code diff and per-video score deltas on the videos both were scored on.
  `{"tool": "compare", "a": "<label or hash>", "b": "<label or hash>"}`.
- `lineage`: the step table of a split record's lineage (every created head and rewrite in step order with its verdict: created, accepted and became the head, or rejected and why; F1, Prec, M%, matched, predictions, the change made).
  `{"tool": "lineage", "ref": "<label or hash>"}`. A `record` query with `"parts": ["critique"]` returns the critique that produced a rewrite; every head line names the record's verdict.

## Logbook
{{ logbook_view }}
{% if best_record %}
## Best record so far
{{ best_record }}
{% endif %}{% if extra %}
{{ extra }}
{% endif %}
Answer with one ```json block of this exact shape and nothing else outside it:
```json
{"queries": [{"tool": "record", "ref": "g1·<objective>·i2", "parts": ["code", "detail"]}, {"tool": "video_history", "video_id": "<video id>"}]}
```
\end{lstlisting}
\end{minipage}\par\medskip

\noindent\begin{minipage}{\linewidth}
\noindent\textbf{Repair.}
\begin{lstlisting}[style=prompt]
# Repair request for objective `{{ objective.id }}`

The candidate you just wrote failed at stage `{{ failure_stage }}` and could not be scored. Error:
```
{{ error_text }}
```

Failed candidate:
```python
{{ candidate_code }}
```

Return a corrected, complete module with the same intent. Keep the reasoning section short (one or two sentences on what was wrong). Remember the rules: only {{ allowed_imports }} may be imported, `predict(video, api)` must exist, return a list of segment/reason dicts, and finish within {{ per_video_timeout_s }} seconds per video.

{{ protocol }}
\end{lstlisting}
\end{minipage}\par\medskip

\noindent\begin{minipage}{\linewidth}
\noindent\textbf{Answer protocol.}
\begin{lstlisting}[style=prompt]
Answer protocol for candidate-producing calls. Your answer must contain exactly these {% if summary is defined and summary %}four{% else %}three{% endif %} markers, in this order:

<<<REASONING>>>
Your reflection: what you concluded from the evidence, what you change relative to the previous candidate and why, and what you expect to improve. Plain text, as long as needed.
{% if summary is defined and summary %}<<<SUMMARY>>>
What the program is, in about {{ summary_words }} words (two lines): its stages, the VLM calls and tools it uses, its key parameters and how it decides what to report. This is what later calls see of the program when its code is not shown.
{% endif %}<<<CANDIDATE>>>
```python
# one complete, self-contained Python module
def predict(video, api) -> list[dict]:
    ...
```
<<<END>>>

Rules for the module: it defines `predict(video, api)` and returns a list of `{"segment": [start_s, end_s], "reason": "<text>"}` dicts (an empty list means no error found); it may only import {{ allowed_imports }}; it must be deterministic (temperature is 0 everywhere) and must finish within {{ per_video_timeout_s }} seconds per video; it never reads files, the network or environment variables; helper functions and constants are allowed; comments explaining the strategy are welcome. Exactly one fenced python block inside the CANDIDATE section.
\end{lstlisting}
\end{minipage}\par\medskip

\end{document}